\documentclass[letterpaper, 10 pt, conference]{ieeeconf}  % Comment this line out if you need a4paper

\IEEEoverridecommandlockouts                              % This command is only needed if 
\usepackage{url}
\usepackage{cite}
\usepackage{xcolor}
\usepackage{graphicx}
\usepackage{wrapfig}
\usepackage[ruled,vlined]{algorithm2e}
\usepackage{algorithmic}

\usepackage{amsmath}
\usepackage{amssymb}
\usepackage{amsopn}
\usepackage{multirow}
\usepackage{rotating}

\usepackage{color}
\usepackage[normalem]{ulem}
\usepackage{listings}

\usepackage{subfigure}

\usepackage[colorlinks=false,pdfborder={0 1 1},letterpaper=true,pagebackref=true]{hyperref}
\usepackage{indentfirst}
\usepackage{grffile}
\usepackage[utf8]{inputenc}

\usepackage[normalem]{ulem}
\usepackage{breqn}

\newcommand{\fa}[1]{#1}
\newcommand{\fd}[1]{}

\title{\LARGE \bf Improving Human-Robot Collaboration via Computational Design}

\author{Jixuan Zhi$^{1}$ and Jyh-Ming Lien$^{1}$%
\thanks{$^{1}$Jixuan Zhi and Jyh-Ming Lien are with the Department of Computer Science, George Mason University, 4400, University Drive MSN 4A5, Fairfax, VA 22030 USA,
        {\tt\small {\{jzhi,jmlien\}@gmu.edu}}}%
        }

\begin{document}

\maketitle
\thispagestyle{empty}
\pagestyle{empty}

\begin{abstract}

When robots entered our day-to-day life, the shared space surrounding humans and robots is critical for effective Human-Robot collaboration. The design of shared space should satisfy humans' preferences and robots' efficiency. This work uses kitchen design as an example to illustrate the importance of good space design in facilitating such collaboration. Given the kitchen boundary,  counters, and recipes, the proposed method computes the optimal placement of counters that meet the requirement of kitchen design rules and improve Human-Robot collaboration. The key technical challenge is that the optimization method usually evaluates thousands of designs and the computational cost of motion planning, which is part of the evaluation function, is expensive. We use a decentralized motion planner that can solve multi-agent motion planning efficiently. Our results indicate that optimized kitchen designs can provide noticeable performance improvement to Human-Robot collaboration.

%When the Human-Robot interactions become ubiquitous, the environment surrounding these interactions will have significant impact on the safety and comfort of the human and the effectiveness and efficiency of the robot. 
%Although most robots are designed to work in the spaces created for humans, many environments, such as living rooms and offices, can be and should be redesigned to enhance and improve Human-Robot collaboration and interactions.
%This work uses autonomous wheelchair as an example and investigates the computational design in the Human-Robot coexistence spaces. 

%Given the room size and the objects $O$ in the room, the proposed framework computes the optimal layouts of $O$ that satisfy both human preferences and navigation constraints of the wheelchair.
%The key enabling technique is a motion planner that can efficiently evaluate hundreds of similar motion planning problems.
%Our implementation shows that the proposed framework can produce a design around three to five minutes on average comparing to 10 to 20 minutes without the proposed motion planner.
%Our results also show that the proposed method produces reasonable designs even for tight spaces and for users with different preferences. 
%\todo{say a few words about the qualify of the design}
\end{abstract}

\section{Introduction}

%When the Human-Robot interactions become ubiquitous, the environment surrounding these interactions will have significant impact on the safety and comfort of the human and the effectiveness and efficiency of the robot. 

When Human-Robot collaborations become more and more prevalent, the  space shared by humans and robots is important for effective and safe interaction. Although most robots are developed for humans in the existing environments, we foresee that the common space is going to evolve in the coming decades as humans and robots work closer together. This paper discusses a space design problem that enhances Human-Robot collaboration. 

%More than ever, robots are designed and developed to work with and around humans. Inevitably, humans and robots in their day-to-day life are going to share a common space.  While most robots, in particular humanoid robots, are designed to work in the existing environments designed for human activities, we envision that the shared space are likely to evolve in the near future to better accommodate and enhance the ever increasing Human-Robot interaction and collaboration. 
%During the decades after personal vehicles were invented, we redesigned the roads and streets to adjust to the vehicle size, speed, traffic volume and, more importantly, the behaviors of the drivers in these cars. 
%Similarly, as robots are moved from industrial and laboratory settings into our personal spaces, the spaces must also adopt to the robots to increase the humans' safety and comfort and the robot's efficiency.

%We propose a computational space design for Human-Robot collaboration. 

Our previous work showed an optimization design of space shared by humans and robots~\cite{zhi2021designing}, however, this optimization framework only considers the human's preference and robot motion. It does not require interaction or collaboration between humans and robots. In this paper, we extend our work considering Human-Robot collaboration on tasks that can be broken down into sub-tasks that require humans and robots to work together to finish the main task.

\begin{figure}[]

{\includegraphics[width = 0.24\textwidth]{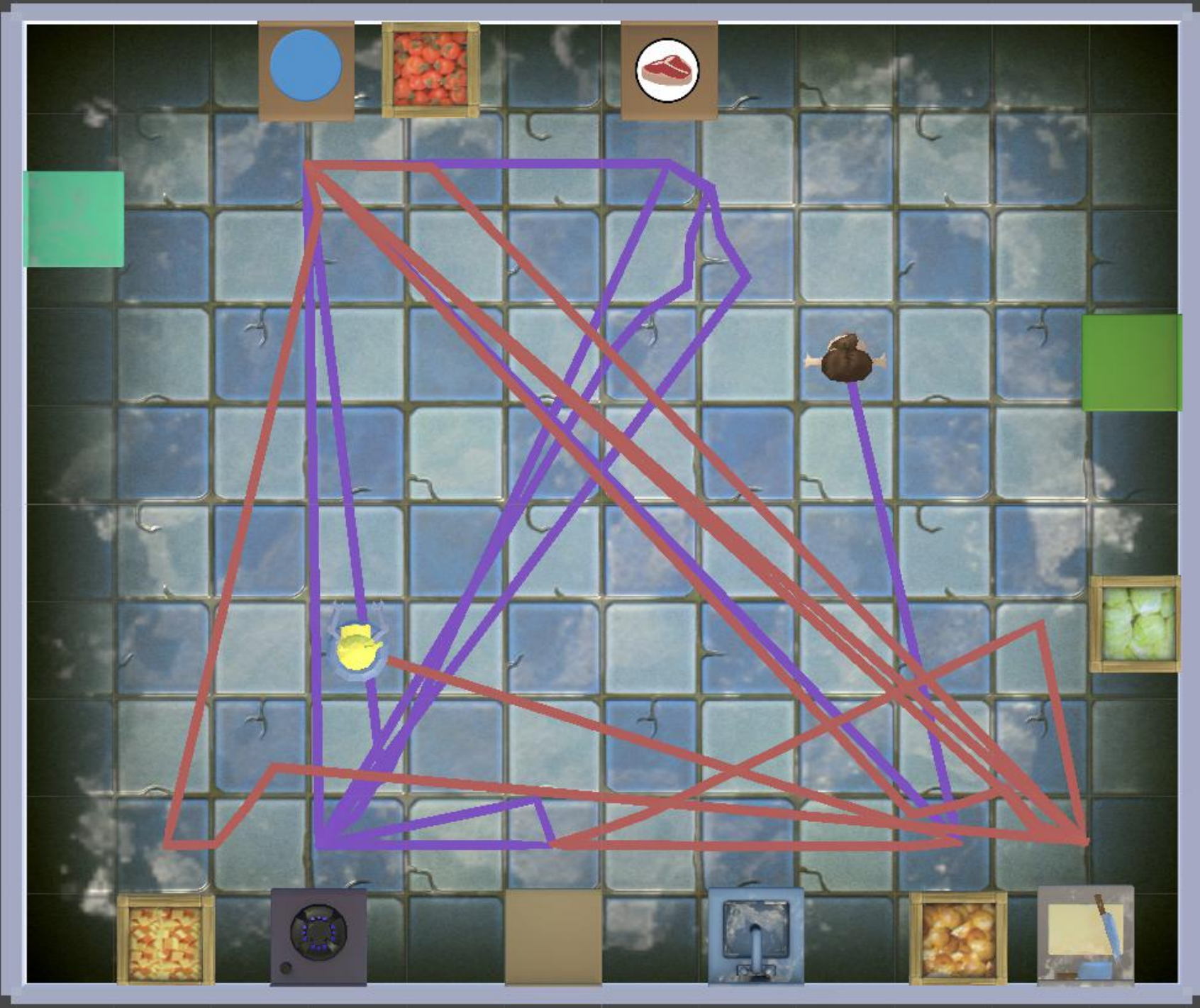}}
{\includegraphics[width = 0.24\textwidth]{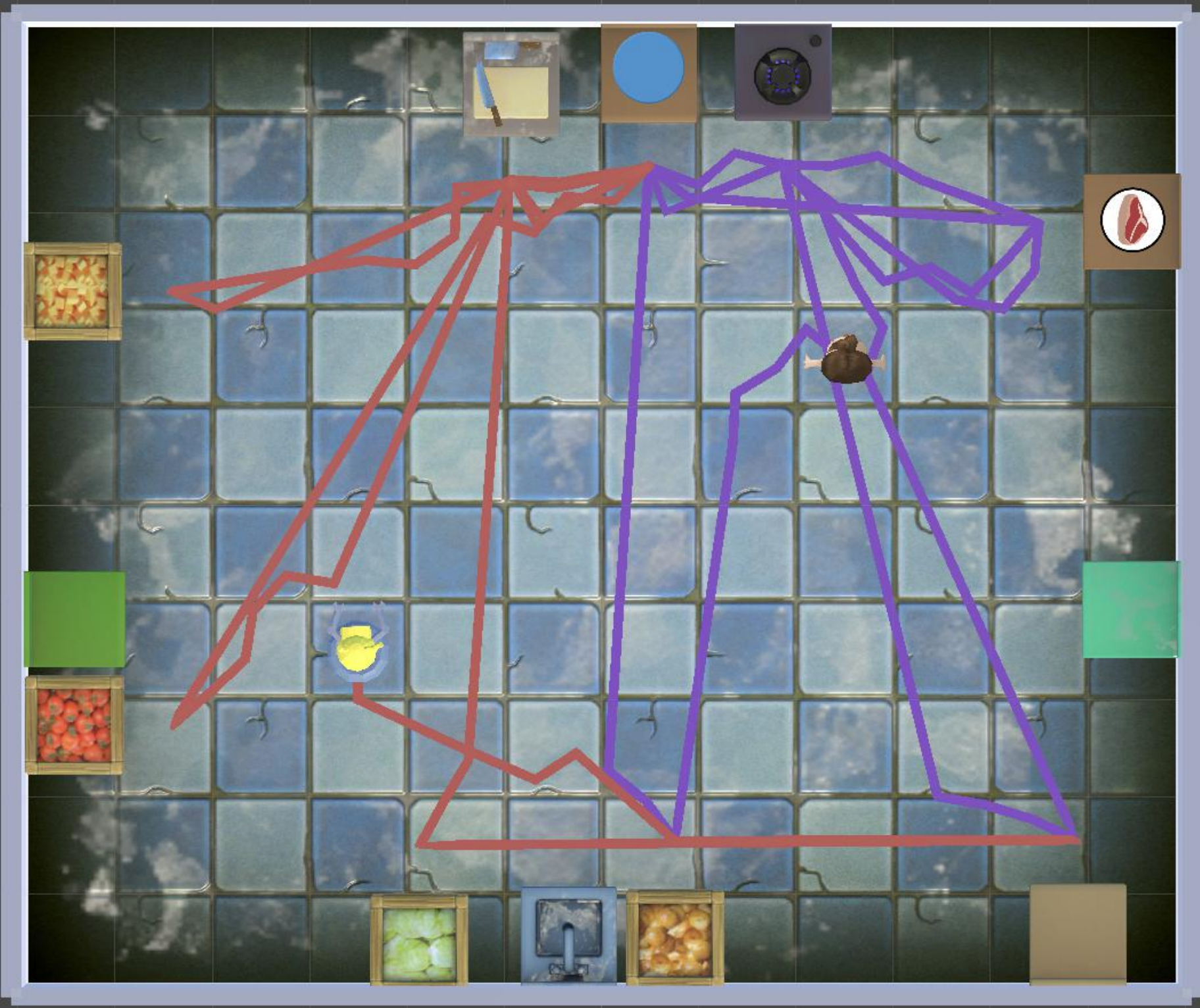}}
\caption{The left kitchen layout leads to worse performance than the right kitchen layout generated by the proposed method when the human and the robot collaborate to make two burgers. The red and purple lines represent the paths of the robot and the human. The paths' lengths in the left layout are longer compared to the paths' lengths in the right layout for both humans and robots. The task completion time of the left layout is also longer than the completion time of the right layout when we calculate the time.   }

\label{fig:kit-comp}
\end{figure}

Inspired by our day-to-day chores, making dishes requires humans to work together. Similar to the gameplay in Overcooked~\cite{ghost2016}, we will use the scenarios that require close collaboration to evaluate the proposed ideas. More specifically, cooking tasks consist of organized sub-tasks,  and humans and robots can work in parallel for different sub-tasks. Additionally, humans and robots should avoid collisions at any time. These coordination challenges are key factors in Human-Robot collaboration.
From another perspective, a better design of the kitchen can improve collaboration in these cooking tasks. These improvements can be measured by the travel distance, path intersection of the human and the robot, and task completion time. Fig.~\ref{fig:kit-comp} presents a comparison between two kitchens designed with and without Human-Robot collaboration taken into account.

%For the robot,  it needs to know what humans are doing and their intentions, then it can plan its own actions to accomplish the task. Therefore, the robot needs to understand the environment, plan a path that does not collide with humans, and accomplish the task. 

%Our previous work proposed the first computational method considering the optimization of Human-Robot coexistence space, which subjects to constraints from both human preferences and robot motion limitations. 

%In this paper, We propose to extend our previous work by adding Human-Robot collaboration for space design optimization. Inspired by Overcooked~\cite{ghost2016}, the kitchen environment is well-suited because it has cooking tasks that involve Human-Robot coordination challenges such as divide and conquer and path coordination. Therefore, We focus on a kitchen space design problem that considers Human-Robot collaboration.

The kitchen design problem for Human-Robot collaboration can be formulated as an optimization problem. In the optimization framework, the decision variables are the layout of counters, and the quality of the paths of humans and robots can be used to measure collaboration. Therefore, the objective function encodes layout information and path quality. The key technical challenge here is the optimization method calls the motion planner thousands of times during the optimization process. Therefore, it's important to solve many motion planning problems first.

%From our study, motion planning takes more than 95\% of the computation time.

%\textbf{Main contributions}. This paper presents the first computational method considering the optimization of the space shared by human and robot.
%The paper also contributes the first nonholonomic motion planner \cite{dolgov2008practical} that adopts the solutions obtained from earlier motion planning problems to solve more problems in similar but new workspaces.
%We envision that, as in many creative processes, the users of the proposed software framework are likely to consider different preferences, objects, room sizes and types, etc, and therefore likely to require multiple optimizations before they can settle on a design. Consequently, an ultra-fast motion planner, such as the one proposed in this paper, is much needed to make this creative process more practical.

%More specifically, the proposed new planner can efficiently plan wheelchair motions in hundreds to thousands similar environments in just a few minutes. 

%\textbf
The main contribution of this work is an optimization framework of kitchen spaces designed for Human-Robot collaboration. The proposed framework can generate kitchen designs that enhance Human-Robot collaboration in a reasonable amount of time. 
%and (2) an effective motion planner that can coordinate human and robot motion by probability inference and reactive planning when the robot does not know the human's task or path information. 

\begin{figure*}[t]
\centering
{\includegraphics[width=1\textwidth]{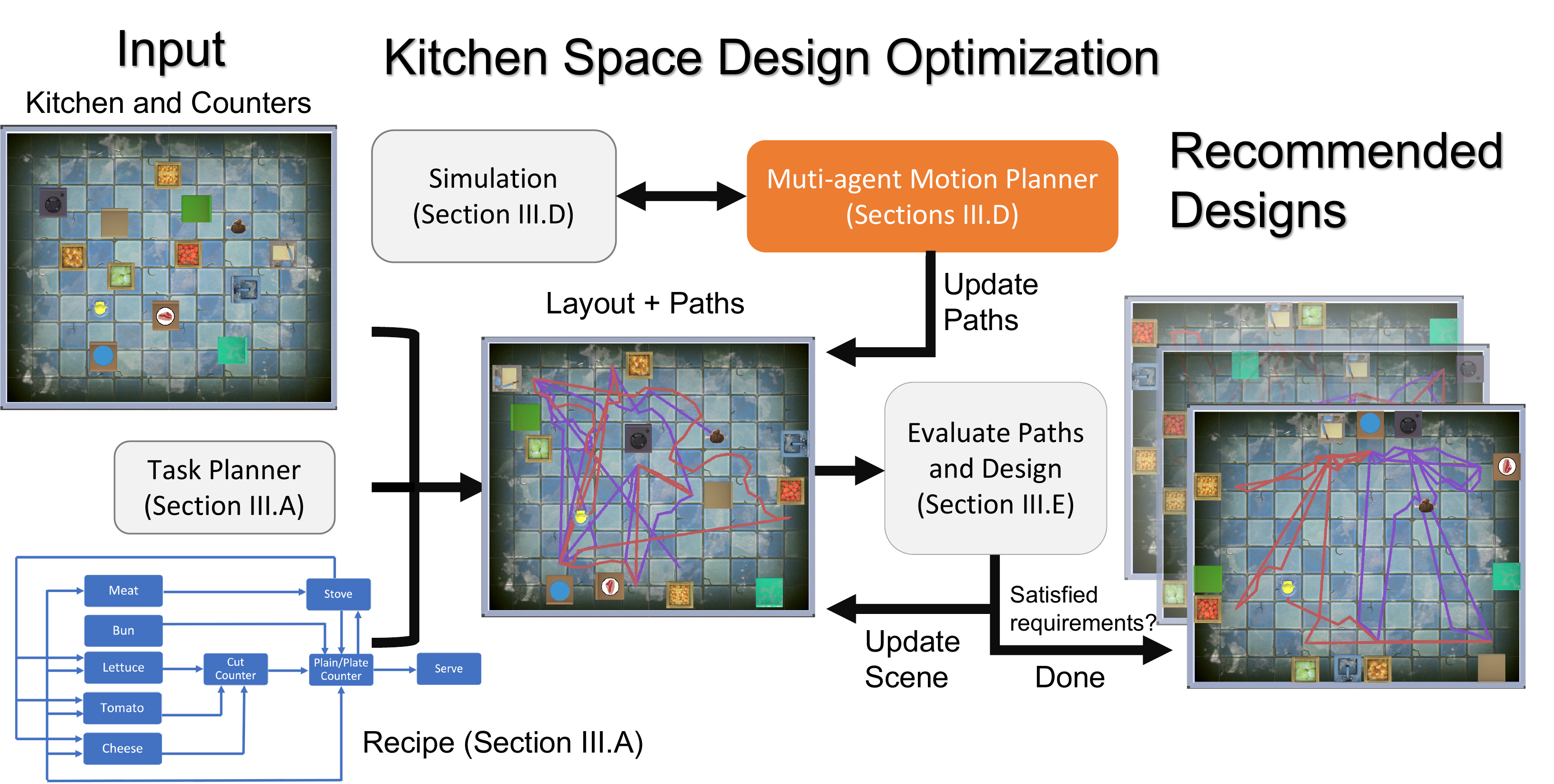}}

\caption{The overview of the kitchen space design. Given the recipes, our method generates kitchens and paths that enhance the Human-Robot collaboration in cooking. }

\label{fig:kitchen-overview}
\end{figure*}

\section{Related work}
Most literature focuses on the computation design of different spaces. However, there is little effort in designing Human-Robot collaboration space~\cite{zhi2021designing}. Several researchers work on Human-Robot collaboration in the existing environment. In addition, Human-Robot collaboration needs coordination and planning for humans and robots.

%There has been extensive work on computational design of living spaces. However, to our best knowledge, there is little effort, if not none, on designing Human-Robot coexistence spaces.

%The other major technical challenge in designing a Human-Robot coexistence space is an ultra fast planner that solves motion planning problems in many similar environments. In this section, we will review prior works in both areas.  

\subsection{Computational Method for Space Design}

%\todo{talk about methods in computational design of spaces; cite the VR wheel chair work.}
Researchers formulated the space design as optimization problems. 
%We group these works in rule-based and activity-based methods. 
%In this work, we use both rule-based and activity-based representations for human preferences and robot motions, respectively. 
Several methods defined different rules to indicate the choices of object positions and orientations. For instance, Yu et al.~\cite{yu2011make} developed an optimization framework to change the placement of objects to meet user-defined requirements. %Xu et al.~\cite{xu2015wall} proposed a new structure named {\em wall grids}, which combined the categories and locations of the objects, to generate indoor scenes. 
%Recently,~\cite{wang2018deep} learn these rules by neural network and use them to create complicated scenes.
Zhang et al.~\cite{zhang2021joint} proposed a computational method that can design a workspace and a workplan together, and applied the method to generate workspaces and workplans in various environments.

Some researchers focus on indoor scene synthesis with human-centric inference. Ma et al.\cite{ma2016action} trained the action graph from annotated photos, which indicates the transition probabilities between the learned actions. Combined with the initial scene, they sampled a sequence of actions from the action graph, each action triggers a change in the scene, %such as adding a new object or arranging a new placement, 
finally they create the optimized scenes. Fu et al.\cite{fu2017adaptive} learned activity-associated object relation graphs for a different type of scene from the floor plan database, which indicates object categories and combinations, then explores the graphs to generate reasonable scenes.

In addition, Some works discuss the relationship between performance and workspace design. Kämpf-Dern and Konkol~\cite{kampf2017performance} proposed a framework for workspace design that considers actors and performance. Ribino et al.~\cite{ribino2018agent} applied agent-based simulations for improving performance in warehouses.

%Human activities have a strong effect on arrangement of objects. 
%For example, Fisher et al.\cite{fisher2015activity} learned an activity model from a database of 3D models and scenes and, based on the activity model, new scenes are generated.
%Several activity models have been proposed, such as a action graph \cite{ma2016action},  activity-associated object relation graphs \cite{fu2017adaptive}.
%  In this paper, an activity model is used to infer the how the wheelchair robot may move around a room.
%. Their activity model consists of two parts: an occurrence model which decides whether an object presents in the activity; an interaction model which determines the arrangement of the objects for the interaction. They also proposed a scene template that involved activity map \cite{savva2014scenegrok} and geometric representation. Combined activity model and scene template, they iterative optimize the 3D scene.

%
%
%Qi et al.\cite{qi2018human} proposed a probabilistic grammar model named And-Or graph (S-AOG) to represent scenes, which learned from indoor scene dataset. In this model, the terminal nodes represent different objects such room, furniture.  Contextual relations on the terminal nodes are human activity. They can sample new reasonable layouts with the model. 

\subsection{Human-Robot Collaboration in the Existing Environments }

Bauer et al.~\cite{bauer2008human} proposed a framework for Human-Robot collaboration, which consists of perception, intention estimation, joint intention, action planning, and joint action. Alami et al.~\cite{alami2005task} proposed a decision-making framework for robot control in a human environment, this framework ensures a robot performs its task while considering human needs and preferences.

Inspired by the popular game Overcooked, some researchers study Human-Robot collaboration in the cooking environment. For instance,  
Carroll et al.~\cite{carroll2019utility} trained the agents by self-play and human data in a simple kitchen environment. The method needs training for different environments. 

%Different models must be trained for different types of tasks and environments.  

%In addition, Wu et al.~\cite{wu2021too} developed Bayesian Delegation that enables robots to infer other members' intentions by inverse planning, then solve the multi-agent cooking challenges that need coordination.
%My work is also related to the cooking environment. Compared to~\cite{carroll2019utility}~\cite{wang2020too}, we study how space design can affect Human-Robot collaboration.  

%Ribino et al.~\cite{ribino2018agent} applied agent-based simulations for improving performance on warehouses. 

\subsection{Multi-Agent Communication and Motion Planning }
Human-Robot collaboration needs communication and planning between humans and robots, here we refer to the literature on multi-agent communication and motion planning. Multi-agent communication has two types: explicit communication and implicit communication~\cite{balch1994communication}, which refer to conveying the message directly or indirectly. For implicit communication, the agent needs to infer another agent's intention by observing their actions. This concept can be applied in Human-Robot collaboration that robots can infer humans' intentions by observing their actions~\cite{wu2021too}~\cite{shum2019theory}.

For Multi-agent motion planning, some researchers proposed a centralized planner that plans the motion of the entire agents simultaneously, but computational time increases exponentially; then decentralized planning is more practical, such as prioritized planning~\cite{van2005prioritized}. The agent is considered in priority order, for instance, the agent with higher priority is planned first, then when considering the other agent, the first agent can be considered as a moving obstacle.

\section{Optimization Framework of Kitchen Space Design}

The components of the proposed optimization framework are introduced in this section. Fig.~\ref{fig:kitchen-overview} presents the optimization framework that uses the designed functionality of the kitchen to analyze the layout of the design. The functionality is defined by a set of dishes/meals provided in the kitchen. Given a list of meals that we wish the robot and the human to collaborate to produce, a task planner will decide the sub-tasks for the human and the robot during the optimization step, then a multi-agent motion planner can generate paths for each sub-task,
a simulator is applied to validate the paths of the human and robot generated by the motion planner.
the framework evaluates the quality of the design by
a cost function encoding the layout information and path quality which measures how the robot and the human collaborate.

%This section introduces the building blocks used in the proposed design framework. 
%Fig.~\ref{fig:overview} illustrates the proposed framework that uses a motion planner and human preferences to evaluate the space layouts.
%The framework also uses an action-object relation graph \cite{ma2016action,fu2017adaptive} that encodes the interactions between the human in the wheelchair  and the objects  in the room. 
%These interactions will determine the importance of a given trajectory and how much the trajectory should influence the design of the room layout. 
%To simplify our discussion, we will use wheelchair to refer to the human in the wheelchair robot for the rest of this paper when the context is clear.
%\todo{we need a figure to illustrate the overall framework. A framework that uses motion planner in the loop to generate space layout. 

\subsection{Recipe and Task Planning}

\begin{figure}[]
\centering
{\includegraphics[width = 0.48\textwidth]{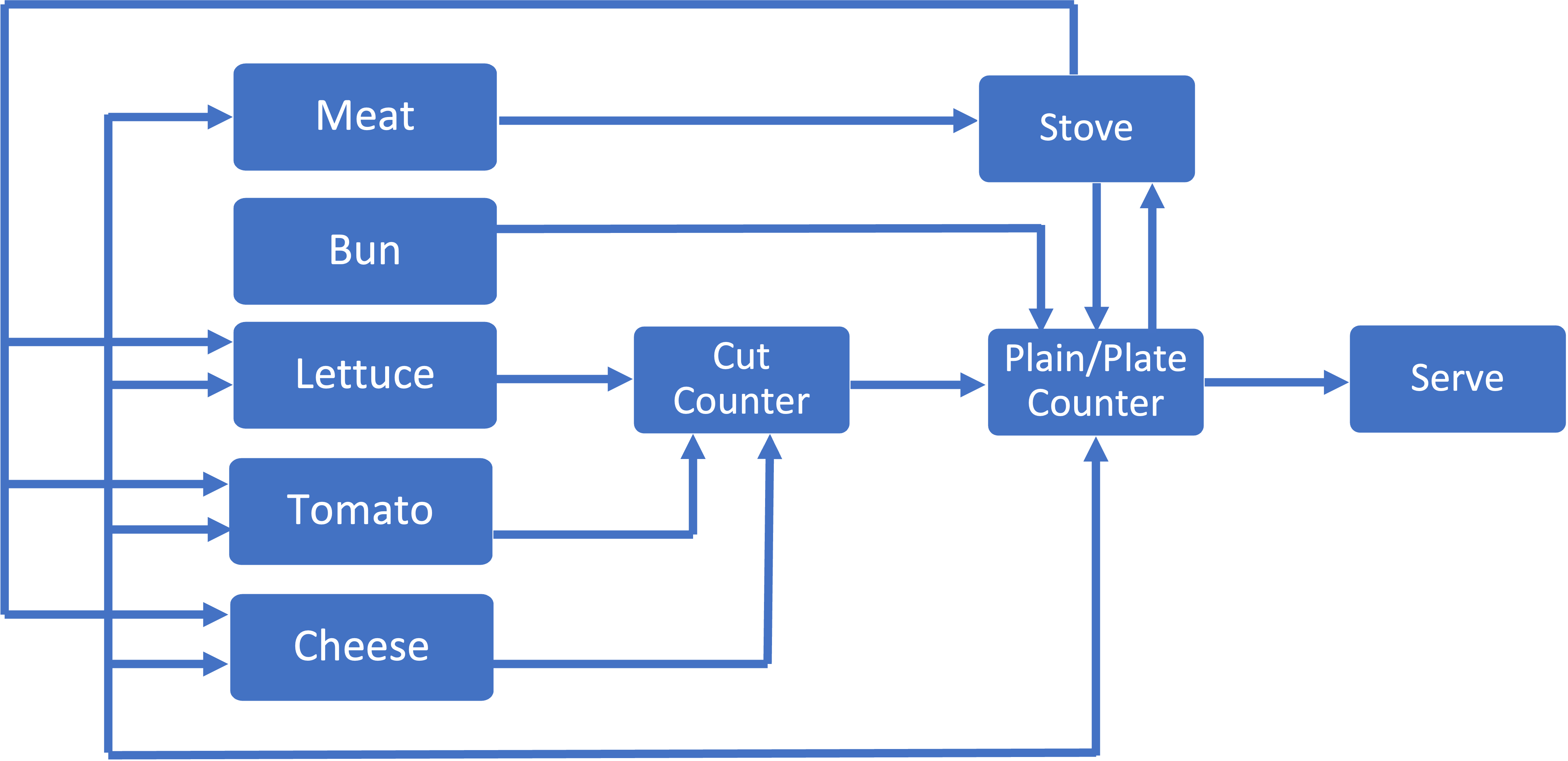}}
\caption{The recipe and sub-tasks to make a burger  }
\label{fig:recipe}
\end{figure}

When humans and robots work together on a common task, we need to assign different sub-task for each partner. For example, to make a hamburger, a bun and meat are required, and cheese, lettuce, and tomato are optional. We create different steps for making a burger, which include preparing the bun, cooking the meat, and adding the cheese, lettuce, and tomato. Each sub-task also needs several steps. For example, a sub-task of cooking meat needs a human or a robot to go to the meat counter to get meat and grill the meat on the stove, then place the cooked meat inside the bun. For sides such as tomato, the human or robot needs to fetch a tomato and slice the tomato then put them in the burger. \fa{ To simplify the work, here we use two burgers as an example to design. Note the work can be extended to a list of dishes.}
The human and robot pick the sub-task alternately and complete the task together.
Fig~\ref{fig:recipe} gives an example of making a whole burger and the counters that need to visit. 

\subsection{Agent Behaviour Design}
\begin{figure}[]
\centering
{\includegraphics[width = 0.48\textwidth]{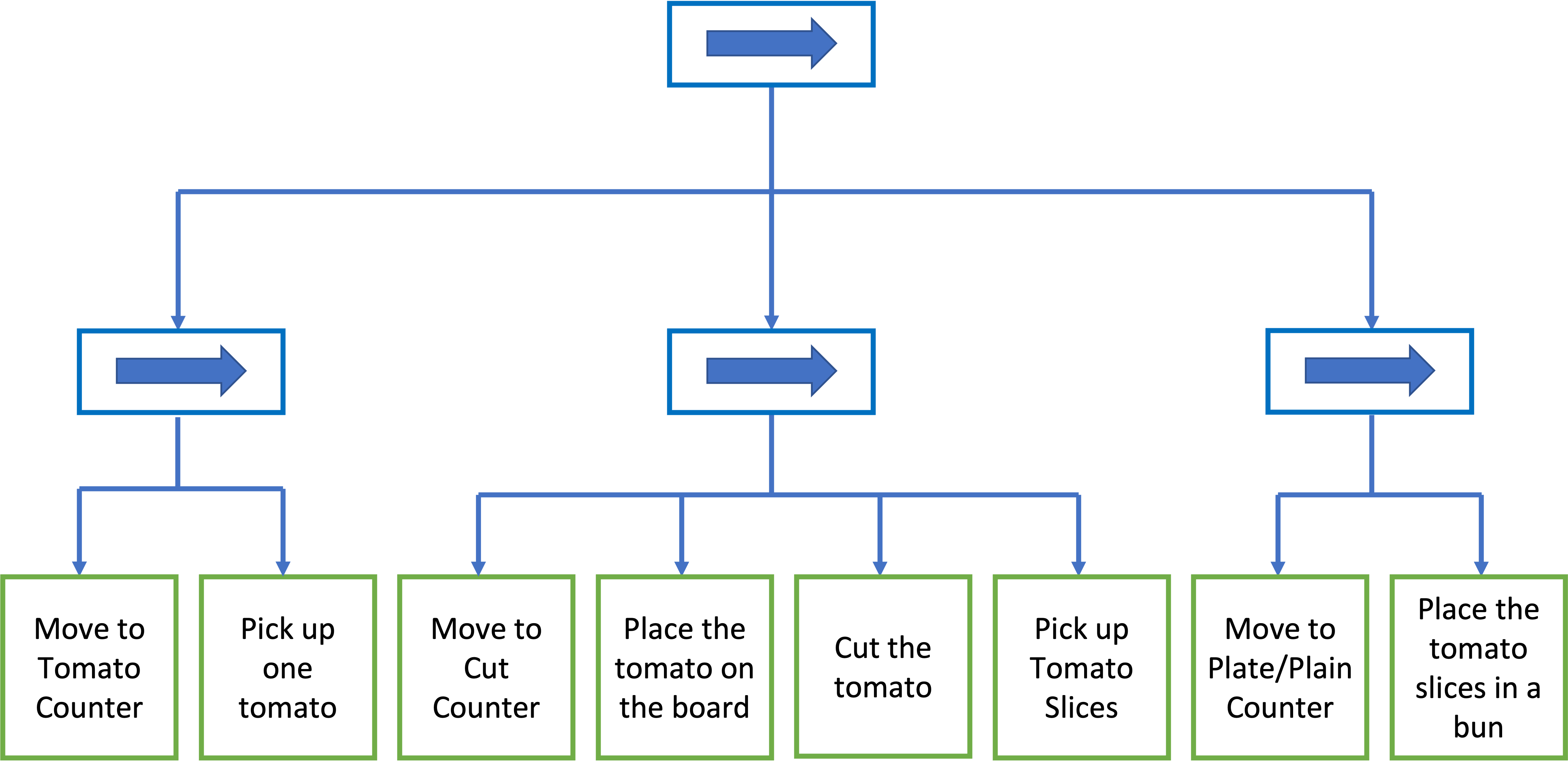}}
\caption{The behavior Tree of a tomato cutting task in making a burger    }
\label{fig:bt}
\end{figure}

To design an agent for task execution, there are two popular techniques: finite state machine and behavior tree~\cite{colledanchise2018behavior}. The finite state machine defines several states for an agent and then defines different actions for state transition. It is simple but not flexible and difficult to modify or extend. A behavior tree is a directed tree structure with different types of nodes. Starting from the root node to the leaf node, the behavior tree can execute different commands to control the agent's behavior. \fa{A behavior tree can also be a sub-tree of a larger behavior tree.} The behavior tree has good modularity and is flexible to modify. However, if the tree is too deep, the behavior tree may need to traverse all the nodes and can be slower. \fa{In this section, a finite state machine and the behavior tree and both used.  A finite state machine is used in a virtual simulator to check the paths of the human and the robot is valid in the motion planning part. See~\ref{sec:mmp}. The behavior tree is used in the real simulator after optimization to show how the robot and the human collaborate in a kitchen. }
Fig.~\ref{fig:bt} shows a behavior tree of a tomato cutting task, robot or human can execute a sequence of actions to perform this task based on the behavior tree.

%When We design the behavior tree, We need to design some conditions to trigger the actions of the robot.  Without communication, the robot can infer the conditions by observing the human position and the information of the task which is encoded in a Bayesian model. To simplify our discussion, the Bayesian model will be given for each cooking task. 
\subsection{Touring the Kitchen}
Cooking with a recipe needs a human or robot to travel to different counters in the kitchen to finish each sub-task. \fa{Consequently, we should find a collision-free path that can visit each counter. Therefore, we need to determine the configuration of each counter for pathfinding.} Then these configurations should be near the counter and should have an orientation so that the person or robot is facing the counter. To simplify the discussion, we only determine one valid configuration related to one counter.

Given the configurations near the counters  $\{c_i,c_j, ..., c_k\}$in a given sequence for each sub-task, we intend to determine a path that covers all counters in an order with no collision. We solve this problem step by step. To find a path between counters $c_i$ and $c_j$, we find the path that connects the pair of each configuration by the motion planner, then the next pair in the order will be processed.  We repeat this process until we find a collision-free path in a given order for a sub-task.

%Note that, in each optimization step, We applied the method in this section to all the sequences obtained from recipes. 

%Recall that, in each layout design optimization iteration, the method described in this section must be applied to all $M=10$ sequences obtained from the AO graph. 
%As detailed in the next section, these $M$ tours will  be used to evaluate the accessibility of the  design.

\begin{figure*}[t]
\centering
{\includegraphics[width=1\textwidth]{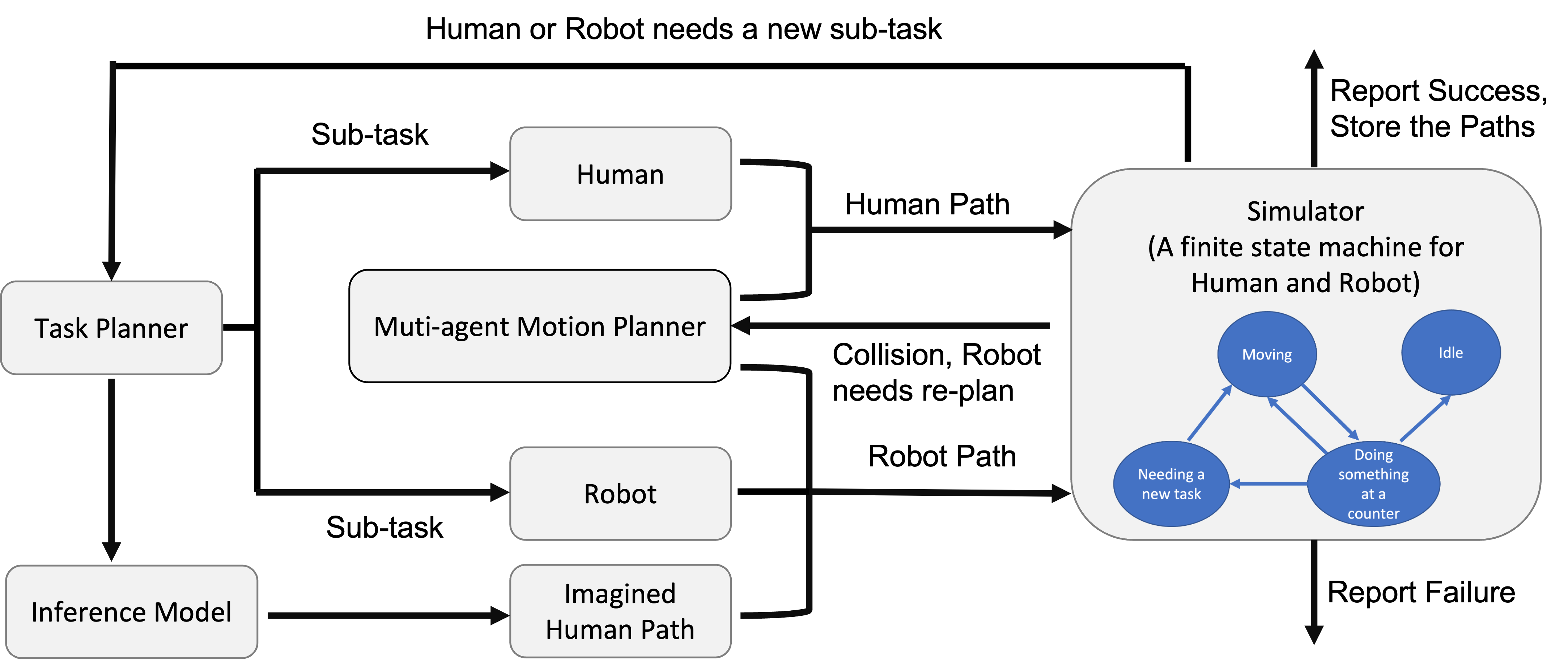}}

\caption{An overview of multi-agent motion planning }

\label{fig:mmp-overview}
\end{figure*}

\subsection{Multi-agent Motion Planning}
\label{sec:mmp}

We now describe the multi-agent motion planner here that
models human and robot as similar agents using Rapidly-exploring Random Trees (RRT) \cite{lavalle1998rapidly} search policy. 

%In Section~\ref{sec:reuse-planner}, We will replace the baseline planner with a more efficient planner that can reuse the experiences for similar environments.  

\textbf{Overview.} For multi-agent motion planning,  we apply a decentralized control policy for each agent because it is faster than the centralized method.   Fig.~\ref{fig:mmp-overview} presents the overview of  multi-agent motion planning. \fa{Note the task planner chooses the sub-task for human and robot randomly, then the robot may estimate or image the states and paths of the human when needed.} To be specific, we choose prioritized planning in this work, the human has higher priority than the robot. Therefore, we plan the human path first, then when considering the robot path, the human can be considered as a moving obstacle. We assume that the human and robot have the same speed during the simulation, and we also specify the time for cooking the meat and chopping the lettuce, cheese, or tomato. \fa{We need to run a simulator to coordinate human motion and robot motion.}

\textbf{Four States in Simulation.} In the simulation, we introduce a finite state machine. We define four states for the human and robot: Moving, Doing something at a counter, Needing a new task, and Idle. When the human or robot is in the Needing new task state, they pick a new sub-task randomly from a task planner. As we talked about before, in each sub-task, we need several paths and operations, which correspond to the Moving state and  Doing something at the counter state. When there is no new sub-task for the human or robot, they are in the Idle state. If they are both in the Idle state, it means they successfully complete all sub-tasks in the simulation, and pathfinding in this iteration of optimization is successful, we can store the results for comparison. Note that the pause for simulations has several reasons: needing a new task for the human or robot, needing re-planning for the robot to avoid a collision, and reporting failure when a collision can not be avoided. The details are described as follows.

\textbf{Planning Human Motion.} 
Since human has a higher priority, the task planner first picks up a sub-task for the human, and the motion planner generates a path for the human.  It uses a modified RRT method with a time component, which means the state $x$ can be represented as $x=(q,t)$, where $q$ is the configuration and $t$ is the time component.

This method starts by generating a random configuration with an 80\% probability of choosing the goal configuration. Then the planner finds the nearest node in the tree under Euclidean distance and connects the randomly generated configuration to the nearest node.
If the goal configuration is selected as the sample configuration, the planner adds the entire path segment until finds the first invalid configuration. Otherwise,  the planner adds up to five steps of the path to the tree~\cite{kuffner2000rrt} for search efficiency. 

Secondly, the planner checks whether the resulting segment is collision-free or not, it only adds collision-free segments and nodes to the tree. The planner repeats this process until the search tree reaches the goal configuration.

Note that we also add a timestamp for each node of the tree. The timestamp for the root of the tree is default 0, it might be changed in the global simulator. By expanding the tree, each valid node has a different time based on its expanding steps from the root. 

When the human or robot is near the counter, they may need some time to finish the cooking task, such as chopping the tomato or cooking the meat.
The cost time should also be added to the valid node. This node usually is the new root of the tree to start a path segment. 

In the end, we find a valid path for a human to visit each counter and finish the sub-task.

\textbf{Planning Robot Motion.} 
Now the task planner needs to pick a new sub-task for the robot and the motion planner needs to find a valid path for the sub-task, it utilizes the same RRT method with a time component. Here we need to consider multi-agent coordination, we address the coordination problem by probability inference model~\cite{wu2021too} and reactive planning.

The probability inference model refers to the robot can infer the intentions of the human without communication. We define the probability for all sub-task to be equal at the beginning. By observing the position of the human around the counter, the robot can infer the next step of the human by probability inference model, which means the robot can generate a virtual path with timestamps for the human by inference model, then considering the virtual path as a moving obstacle for the robot, the robot can plan its own path by checking the collision with the virtual human path. 

However, our experiments show that planning robot motion with an estimated human path increase 20\% time cost, but the path quality does not improve too much but is even sometimes worse compared to planning robot motion directly. This is because the virtual human path is not equal to the real human path, and it affects the robot's path even if the real human path does not have an impact on the robot's path. Therefore, we only consider the virtual path when the simulator reports a collision. 

While planning the robot's motion directly, we need to run simulations to check the collision status of paths for the robot and the human. In the simulator, there is a global time. In addition, we know that the expanding process of RRT has the same step distance for the human and robot, we can easily find the node of the human path segment which has the approximate timestamp with the current robot path node,  then we can check the collision for each node of the path by global time and expanding steps of each node. 

When finding a collision, we pause the simulation. The robot needs a reactive behavior to find a new path to avoid the collision, this re-planning process also applies the RRT method and a virtual human path, then we run the simulation to check the new path. If the planner can not solve the collision problem, the simulator reports failure and stops.

\textbf{Iteration}. When the robot and human finish the current sub-task, we update the inference model by the completed sub-task information which means the inference model decreases the probability of this sub-task and increases other probabilities of this sub-task, the total sum of probabilities will still be one. 
Then the task planner will pick up a new sub-task for the human or robot. Since the human has a higher priority, the human does not need to consider the robot's path, and the robot needs to consider the human's path and needs to re-planning if there is a collision in simulation. All the steps are as same as we talked about before until there is no task for assignment or report failure.

\subsection{Evaluating a Space Design}
\label{sec:cost}
%In order to design reasonable room arrangement for robots, our technique approach optimizes robot path and scene arrangement, which are encoded by 
We evaluate the quality of kitchen space by two cost factors. The first one evaluates the status of the placement of the counters. The second one measures the path quality when completing a task.

%\todo{briefly discuss the optimization framework}
%\subsection{Cost terms}
More precisely, we describe the cost function of a kitchen space design as $C_{total}$:
\begin{equation}
C_{total}= \mathbf{C_Lw_L^T} \ + \mathbf{C_Pw_P^T} \ ,
\end{equation}
where $\mathbf{C_L}$ and $\mathbf{C_P}$ are vectors of kitchen layout costs and paths costs, and $\mathbf{w_L}$   and $\mathbf{w_P}$ represent vectors of weights. 
While $\mathbf{w_L}$ is determined by user-defined preferences  
Because of the recipes and task planning,  multiple paths for the human and robot are considered. Therefore,  the total path cost  is computed by adding up the costs of all paths with the same weight. A smaller $C_{total}$ means a better space design. We describe the details below.

%In our optimization frame work, We consider multiple paths (tours), each is generated a sequence of objects in  the AO graph. 
%Therefore, the overall path cost can be acquired by each path cost $C_s(P)$ for the sequence and the likelihood $L(s)$ defined from the AO Graph for each sequence. 
%$C_{totalpaths} =\sum_{s \in S} L(s) C_s(P)$,

%$C_P^l, C_P^r, C_P^n$ and $C_P^a$ means path considerations: the total length of the path, the total rotation of the path, the narrowness of the path and the available of the path. $C_I$ is a vector of interior design costs encoding distance and rotation cost between objects and $w_I$ means the weights of theses costs. 

\subsubsection{Kitchen layout cost} This cost measures the quality of counters placement. Those placements can be described by the spatial relationship among objects. The spatial information includes position and orientation. 

We assume all the counters are along the wall, the wall includes an invisible wall in the center of the space.

\textbf{ Distance Cost $C_I^d$.}
Each counter in the kitchen has a user-defined distance to its closest wall. We define the cost here:
\begin{equation}
C_L^d = \sum_{\forall {i} \in C}({||c_i-w||-d})^2\ ,
\end{equation}
where $c_i$ is the position of counter $i$, $w$ is the nearest wall position to the $c_i$, and $d_{i}$ refers to the target distance. 

\textbf{ Rotation Cost $C_I^r$.}
We assume each counter only has four directions and should have the same direction as its nearest wall. Thus we define the  rotation cost as follows:

\begin{equation}
C_L^r = \sum_{\forall {i} \in C}({||\theta_i-\theta_w||})^2
\end{equation}
where $\theta_i$ is the orientation of the counter $i$ , the $\theta_w$ is the orientation of the nearest wall to the counter $i$.

\textbf{ Kitchen Layout Cost Weight $\mathbf{w_L}$.}
The weights $\mathbf{w_I}$ are defined as  $\mathbf{w_L}=\alpha[w_L^d, w_L^r]$, where $w_L^d$, $w_L^r$ that reflects the significance of the distance and rotation in this optimization framework. If we don't add distance cost or rotation cost, the counter may place in a random position or orientation.

\subsubsection{Path Cost}
\label{sec:pathcost}

For path cost, we have the formula $\mathbf{C_P}=[C_P^l, C_P^t, C_P^n]$, which measures the length, the task completed time, and the narrowness of the path, respectively.

%A path starts at the door position or any other position around start furniture and ends at the position around target furniture. Since the target furniture is randomly changed during the optimization process, the success path is optimized as the furniture changed.  
%A path’s quality could be determined by its length, angle and narrowness. 
For each path of human and robot, a better path means it is shorter and wider. For each cost term, the cost is a scale from 0 to 1. We give each cost term to be 1 if we can not find the path. 
Suppose there is a path consisting of $N$ nodes and $p_i$ refers to the position of the $i$-th node, then we can define the path cost factors.

\textbf{Path Length Cost $C_P^l$}. For path length, we add up the distances between successive nodes along the path. % with same forward or backward information. 
Given a path with $N$ nodes, we can determine the path length by:
\begin{equation}
C_P^l = \sum_{i=1}^{N-1}{||p_{i+1}-p_i||}\ .
\end{equation}
%The backward part is the same as the forward part. 

\textbf{Path Time Cost $C_P^t$}.
Note that the task completed time is related to the path length, but we only care about the completed time of the dish and do not compute the wash time of the plate because the customer only cares about the finish time. Therefore, the timestamp of the submission counter is the path time cost.
\begin{equation}
C_P^r = T_{finished}\ .
\end{equation}
%Additionally, If the path is not available, the path angle cost is a large number. 

\textbf{Path Narrowness Cost $C_P^n$}.
We use narrowness cost to describe the min narrowness along the path. %It can be computed as the difference between the mean width of the waypoints and the user-defined path width $n_{path}$.
It can be computed as the minimum width of the nodes.

\begin{equation}
C_P^n =  2* \min{(||p_i-q_i||,||p_i-w_i||)}\ . % - n_{path})  }
\end{equation}
For each node in the path with position $p_i$, the orientation of  $p_i$ is facing the next node $p_{i+1}$, then we can find the closest object of the node on the left and right sides. The positions are 
$q_i$ and $w_i$ respectively.

%In addition, if the path is not available, the path narrowness cost is a large number. 

%\textbf{Path Available Cost.}
%In each iteration, for a new layout of the room, We may not find the success path due to probability.  Or the old path may not be available for the new layout, in other words, there exists a collision with other objects for the old path in the new layout. If the path is not available, the cost is 1, otherwise, the cost is 0. 
%\textbf{Combined Cost  of Multiple Paths}
%In our optimization frame work, We consider multiple paths (tours), each is generated a sequence of objects in  the AO graph. 
%Therefore, the overall path cost can be acquired by each path cost $C_s(P)$ for the sequence and the likelihood $L(s)$ defined from the AO Graph for each sequence. 
%$C_{totalpaths} =\sum_{s \in S} L(s) C_s(P)$,

\textbf{Path Cost Weight $\mathbf{w_P}$.} %\todo{still very unclear}
%For each  object $o_i$,  We define the weights $\mathbf{w^i_P}=\alpah_i[w_P^l, w_P^r, w_P^n]$, where $\alpah_i$ is a scaling factor determined based on the Action-Object Relation Graph (AO graph), and 
%$w_P^l$, $w_P^r$, $w_P^n$ are user defined parameters for representing the importance of the length, rotation and clearance. 
We use the weights $\mathbf{w_P}$ to decide the combination of the path cost. The weights can be defined as $\mathbf{w_P}=[w_P^l, w_P^r, w_P^n]$, where $w_P^l$, $w_P^r$, $w_P^n$ are user-defined parameters that represent the importance of the path length, time and narrowness.

\textbf{Multiple Paths Cost.}
We consider multiple paths of the human and robot based on recipes and task planning. The total paths cost are the sum of each path cost with a weight. For path length cost and path time cost, the weight is the same. For path narrowness cost, we consider the worst case because the accident may occur in a corner case. Therefore, we ignore the normal cases where path narrowness costs are zero and average the valid path narrowness costs.

%For instance, given a sequence
%$o_1\rightarrow {o_2} \rightarrow {o_3}$ with the estimated likelihood $p$, the count and frequency of $(o_1,o_2,o_3)$ is $(1,2,1)$ and $(0.25,0.5,0.25)$, respectively.
%Then the scaling factor is simply $\sum_s{p_s(0.25,0.5,0.25)}$. 

\subsection{Design Optimization}
This section describes how the proposed framework designs a kitchen space by minimizing costs. In specific, the simulated annealing algorithm~\cite{kirkpatrick1983optimization} with the Metropolis criterion~\cite{metropolis1953equation} is utilized for the optimization. Firstly, an objective function similar to Boltzmann's function for a current state $(P,L)$ is defined as follows:

\begin{equation}
f(P,L) = exp(-\frac{1}{t}C_{total}(P,L))\ ,
\end{equation}

where $P$ are the set of robot and human paths for the cooking task in the kitchen room. $L$ is the placement of all counters, and $t$ represents the parameter of temperature, which gradually decreases during the optimization. 

The proposed optimization method has two stages, layout optimization, and task paths optimization, those two optimizations are applied in turn.

In specific, in every step, the proposed framework makes a change from the current state  $(P, L)$  to a new state $(P', L)$ or $(P, L')$. We define two types of moves in layout optimization:
(1) change the position and orientation of the counter, and (2) swap two counters. 
For path optimization, the task and path planning will generate new order of tasks and new paths.
Then the framework creates a new design $L'$, and the task and motion planner find a new set of paths $P'$. % based on the new position of target.

The solution space is explored efficiently by the simulated annealing method since the method proposes a large change at first and adjusts the state with a minor change at last.
The proposed framework accepts the new state with a probability derived from the Metropolis criterion:

\begin{equation}
Prob(newConf|oldConf) = min(1 , \frac{f(newConf)}{f(oldConf)})\ .
\end{equation}
When the optimization starts and the temperature is high, the optimizer may accept a bad change. When the temperature decreases, the optimizer acts like a greedy method and is unlikely to accept a bad change. 

\textbf{Alternating Optimization}. Optimizing the layout and task paths simultaneously would be challenging because the trade-offs between the objectives may not be clear, then it is easily stuck in the local minimum.
Instead, the proposed method applies an alternating optimization mode. 

\subsection{Solution Set}

The linear weighted sum method is sensitive to the choice of weighting factors of layout and path cost, and may not always produce a good trade-off between these two objectives. To overcome this limitation, we applied an advanced method called Non-dominated sorting~\cite{srinivas1994muiltiobjective} which sorts a set of solutions relying on their level of dominance. 
One solution is considered to dominate another solution if it is superior in at least one objective and is not inferior in any other objectives. We maintain a solution set with a fixed size (5) called Pareto set to store the solutions. In each iteration, when the framework accepts a solution, we checked the dominance level of the solution in comparison to the stored solutions in the Pareto set. If the newly generated solution dominates one or more solutions stored in the Pareto set, we delete the dominated solutions and put the new solution into the Pareto set. Note when the Pareto set is full, we sort the solutions based on the total path cost and delete the solution with the largest cost. With this Non-dominated sorting, we could find the Pareto front of this space design problem.

\section{Experiments and Result for Kitchen Design}
In this section, we study the quality of the kitchen designs created by the proposed method. We implemented the optimization framework and the multi-agent motion planner in the Unity game engine. 
%The experiments are in a MacBook Pro laptop with Apple M1 Pro and 16 GB memory.

%We demonstrate how our method can be applied in three different room types: bedroom, office, and living room. 
%For each environment, We extract 10 most possible sequences from a given AO graph.
% We set the maximum number of iteration which is 700 to create the graph.
%The maximum number of optimization iterations in our simulated annealing  is 800 for all experiments.

\begin{figure}[htp]

{\includegraphics[width = 0.24\textwidth]{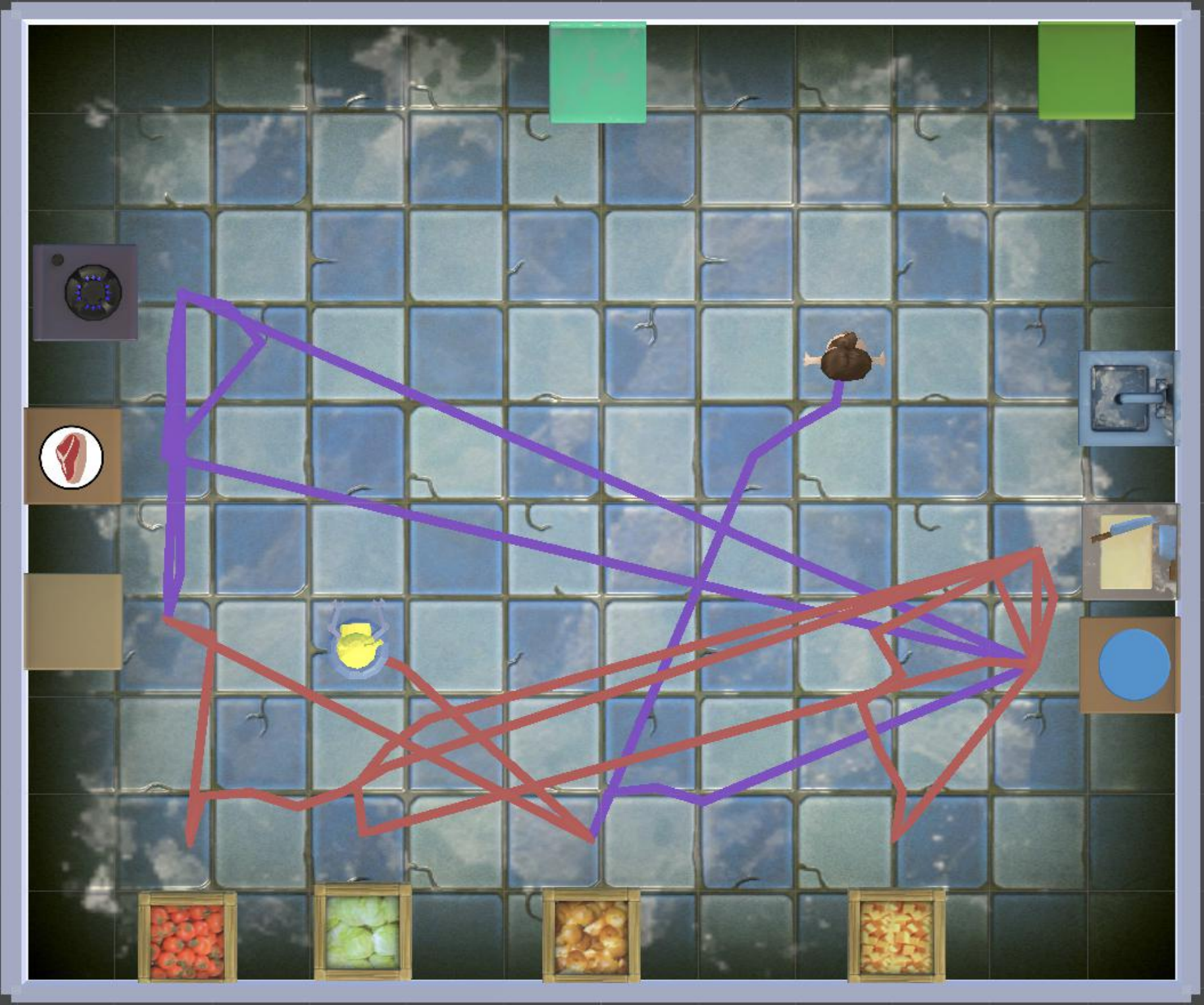}}
{\includegraphics[width = 0.24\textwidth]{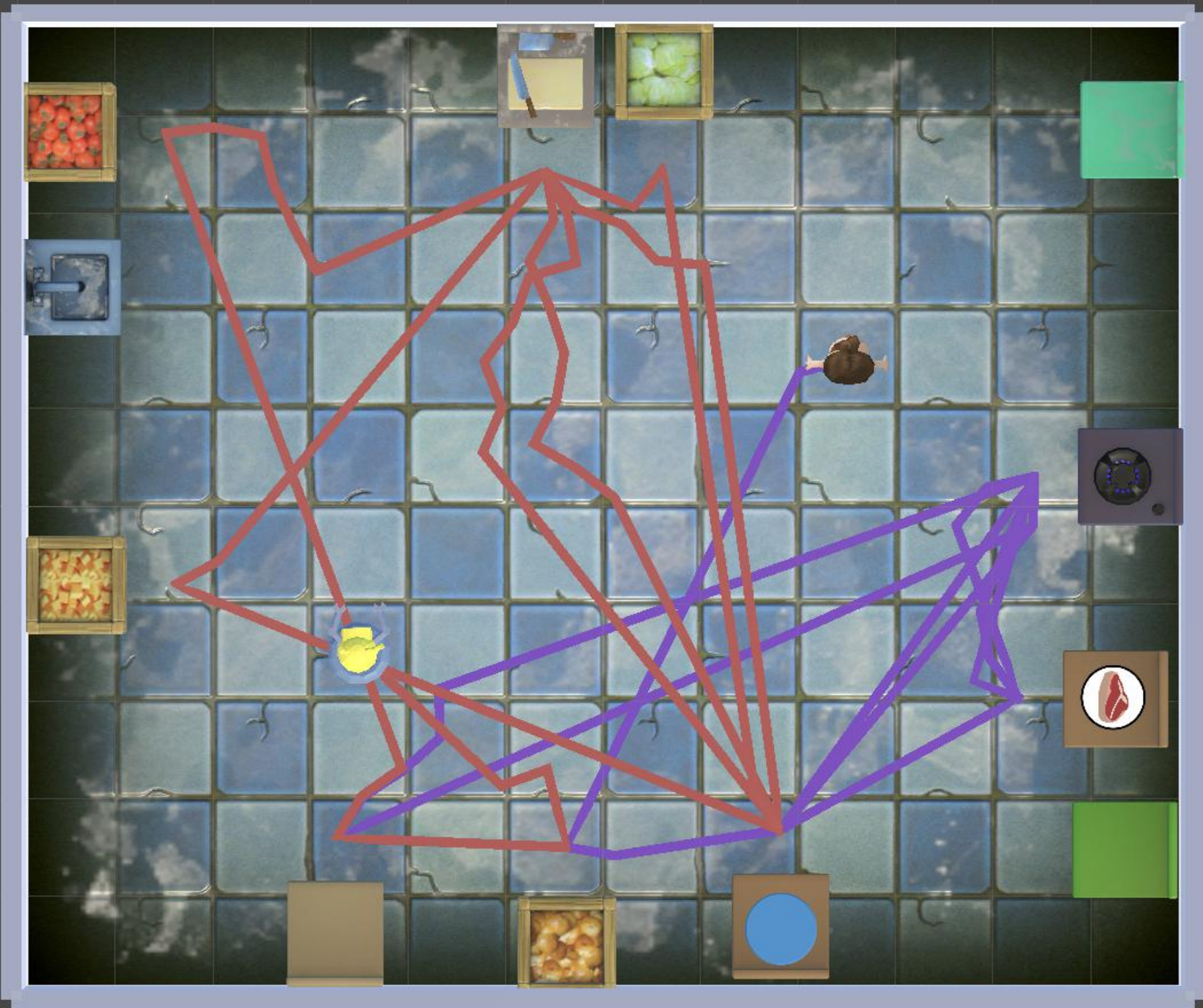}}

{\includegraphics[width = 0.24\textwidth]{Kfigs/regular/81V-min.pdf}}
{\includegraphics[width = 0.24\textwidth]{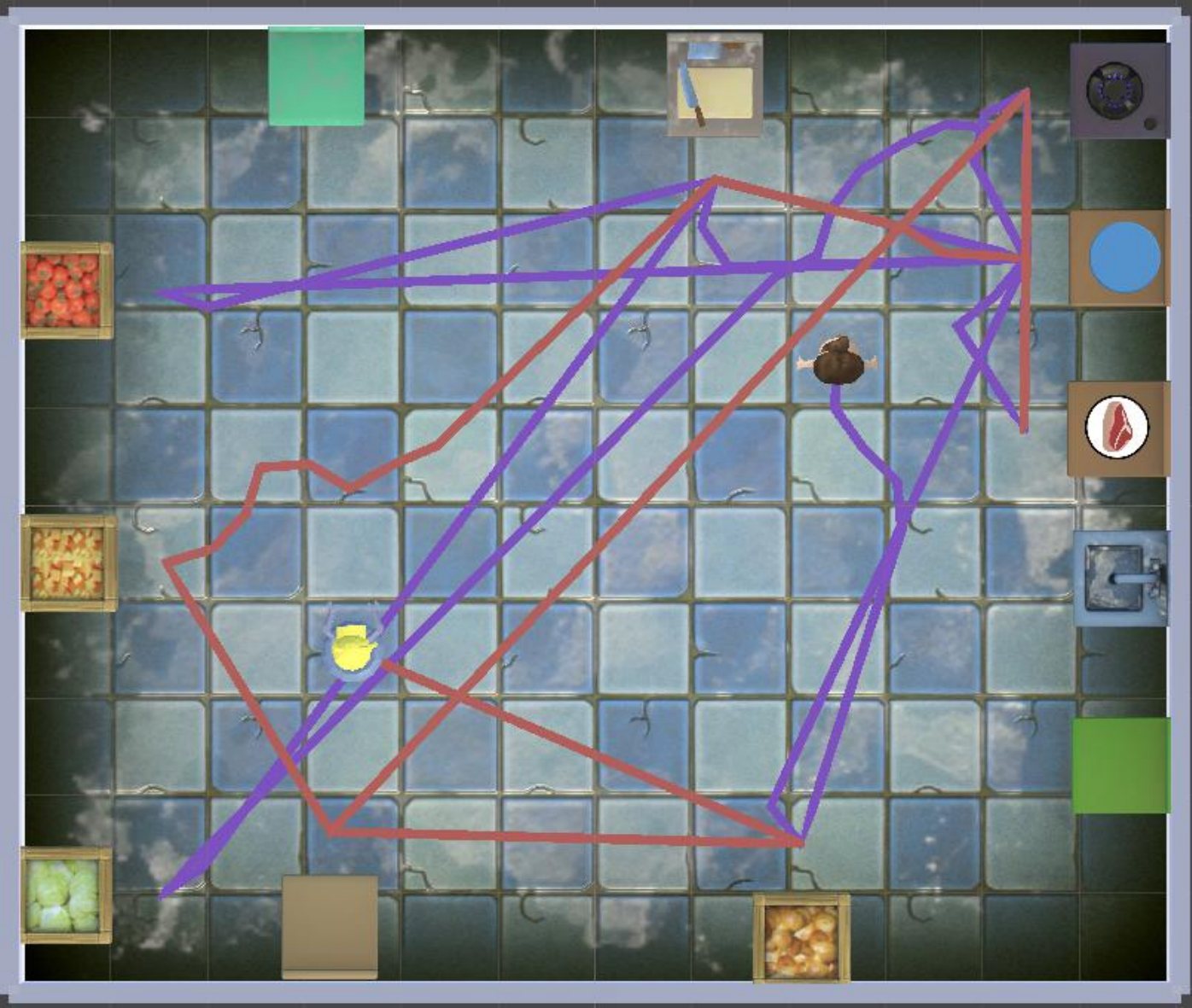}}

\caption{Four kitchen designs created by the proposed method. The red and purple lines represent the paths of the robot and the human. Solution 1 (top left), Solution 2 (top right), Solution 3 (bottom left), Solution 4 (bottom right).}
\label{fig:kitchen}

\end{figure}

\subsection{Results Analysis}
\label{sec:ana}
\begin{figure}[]
\centering
{\includegraphics[width = 0.48\textwidth]{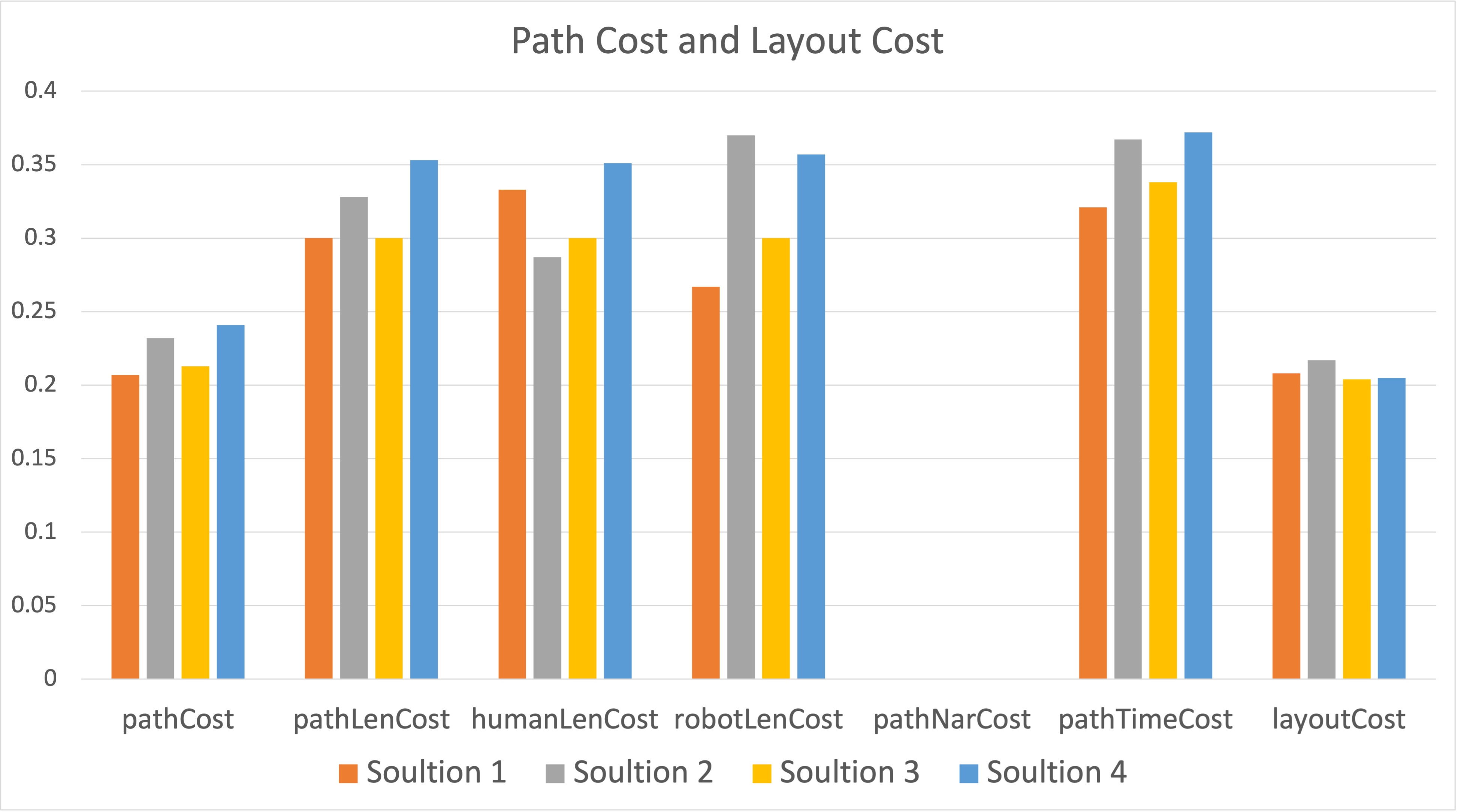}}
\caption{Path costs and layout costs of four designs in Fig. ~\ref{fig:kitchen} }
\label{fig:plcost}
\end{figure}

In this experiment, the human and the robot work together to prepare two burgers. One burger has beef, cheese, lettuce, and tomato; and another burger only has beef. Therefore, we have seven sub-tasks: bun-1, meat-1, tomato-1, lettuce-1, cheese-1, bun-2, and meat-2. Here we define that burger 1 is placed on the Plate counter, and burger 2 is placed on the Plain counter. We repeated the experiments 10 times and got a total of 50 Pareto optimal solutions. Fig~\ref{fig:kitchen} shows four different kitchen layouts. Fig~\ref{fig:plcost} shows the different types of path cost and layout cost in these four different layouts. Table~\ref{tab:task} shows the chosen task planning of the human and the robot by the motion planner during the optimization
The lower path cost of any type indicates higher path quality, thus better Human-Robot collaboration.

Comparing Solution 1 and Solution 2, they have the same task plan for the human and the robot, but Solution 1 prefers a lower robot path length cost, Solution 2 prefers a lower human path length cost, and Solution 3 and Solution 4 are balanced for robot path length cost and human path length cost. However, Solution 4 has a different task plan for the human and the robot. In our results, 72\% of tasks for the human are to place a bun and cook two pieces of beef. It is easier to find Pareto optimal layouts and paths when using this particular type of task plan, that's why Solution 4 has a larger path cost than that of the other three solutions. Note that all solutions have zero Path Narrowness Cost.

\begin{table}
\caption{Task Planning for Each Solution}

\begin{center}

\begin{tabular}{p{1.2cm}||p{2.8cm}|p{3cm}|}
 \hline
\bfseries  &  \bfseries Human Task	 &\bfseries Robot Task\\
  \hline
Solution 1    &	bun-1, meat-1, meat-2		 & bun-2, tomato-1, cheese-1, lettuce-1  \\
  \hline
Solution 2     &	bun-1, meat-1, meat-2	 &  bun-2, tomato-1, cheese-1, lettuce-1 \\
    \hline
Solution 3     &	bun-1, meat-2, meat-1	 & bun-2, lettuce-1, cheese-1, tomato-1 \\
   \hline
Solution 4    &  bun-1, meat-1, lettuce-1, tomato-1	 &bun-2, cheese-1, meat-2\\ 
  \hline
\end{tabular}
\end{center}
\label{tab:task}

\end{table}

\subsection{Optimization Comparison}
The two cost terms,  layout cost and path cost,  are optimized together in our framework. We can also optimize them separately, we first optimize the kitchen space layout until the layout cost is less than a threshold, then perform optimization on task planning and pathfinding in the generated layout with the defined recipes. By comparing the performance of  optimization together and optimization separately,  we can determine which one is better for the space design problem in terms of layout and path quality, and also we can see whether there is a trade-off between layout and path quality. 

Fig.~\ref{fig:kit-comp} presents a comparison between two kitchens with paths. The left figure is the result of optimizing separately, and the right figure is the result of optimizing together. Both kitchens have the same task plan for humans and robots. The right one has a better path quality than the left one which means it is more suitable for Human-Robot Collaboration.

\begin{figure}[]
\centering
{\includegraphics[width = 0.48\textwidth]{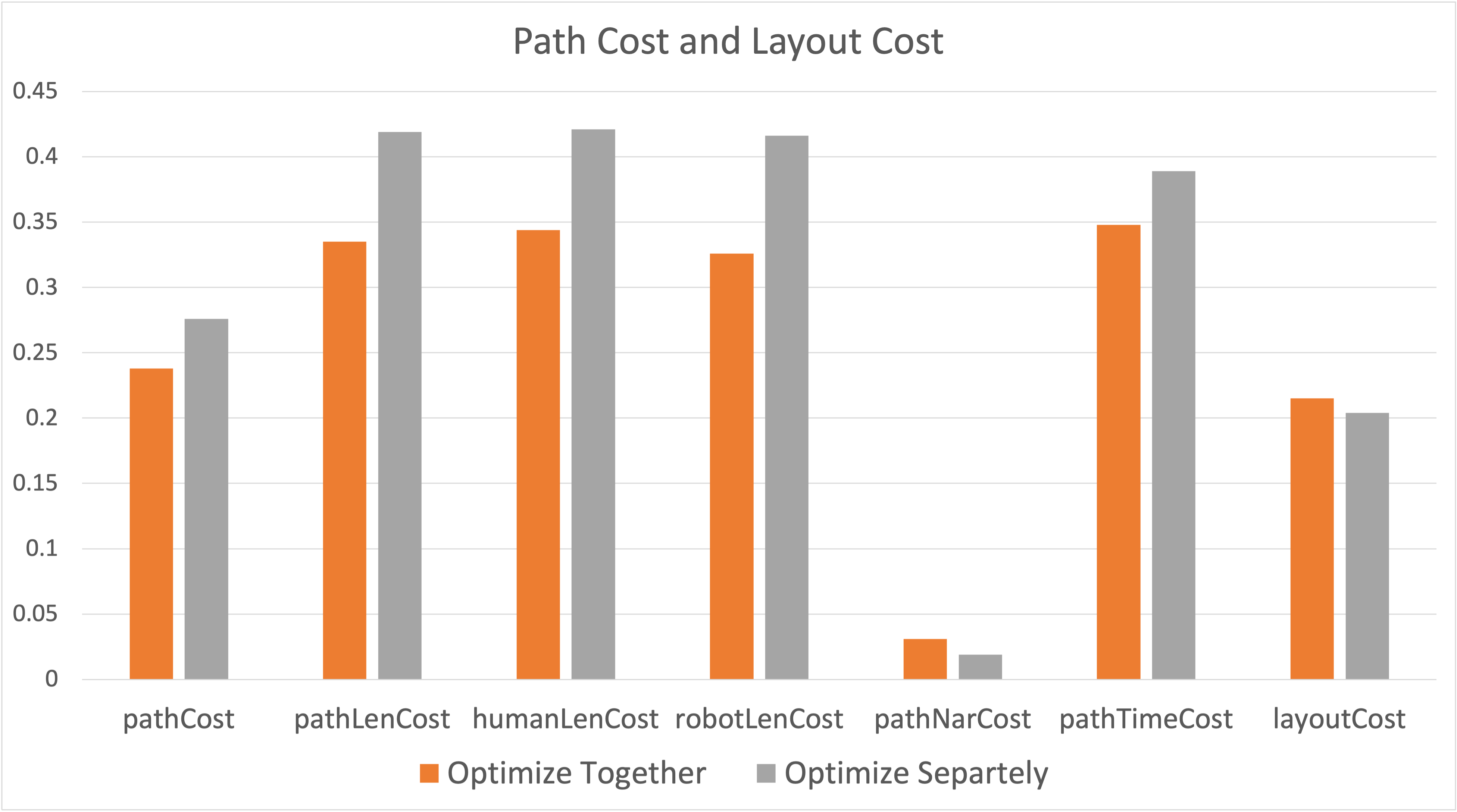}}
\caption{Path costs and layout costs when optimizing together and separately.  }
\label{fig:pc}
\end{figure}

Fig~\ref{fig:pc} shows the different types of path costs and layout costs when optimizing together and optimizing separately. Except for path narrow cost and layout cost, in every other case, the costs of optimizing separately are higher than the costs of optimizing together, which means the path generated by optimizing separately is worse than the path generated by optimizing together. Optimizing together has a better balance between layout cost and path cost. In most cases, the costs are above 10\% higher than the proposed method. This is reasonable because when optimizing separately, some layouts generated by the first step are difficult to yield high-quality paths in the second step. But for the proposed method, optimizing the layout and path together rules out the layouts that can not lead to high-quality paths.

%\begin{figure}[]
%\centering
%{\includegraphics[width = 0.48\textwidth]{Kfigs/pathlength.png}}

%\caption{Different types of path length when optimizing together and %optimizing separately. }
%\label{fig:pl}
%\end{figure}

%\begin{figure}[t]
%\centering
%{\includegraphics[width = 0.48\textwidth]{Kfigs/finishtime.png}}
%\caption{Task completion time when optimizing together and optimizing separately. }
%\label{fig:ft}
%\end{figure}

%We also show the different types of path lengths and finish times without normalization. Fig~\ref{fig:pl} shows the different types of path lengths when optimizing together and optimizing separately. In every case, the path length of optimizing separately is about 20\% higher than the path lengths of optimizing together. %Fig~\ref{fig:ft} shows the task completion time when optimizing together and optimizing separately. It is clear that the finish time of optimizing separately is about 10\% higher than the finish time of the proposed method.

\subsection{Design a Small Kitchen}

\begin{figure}[]

{\includegraphics[width = 0.24\textwidth]{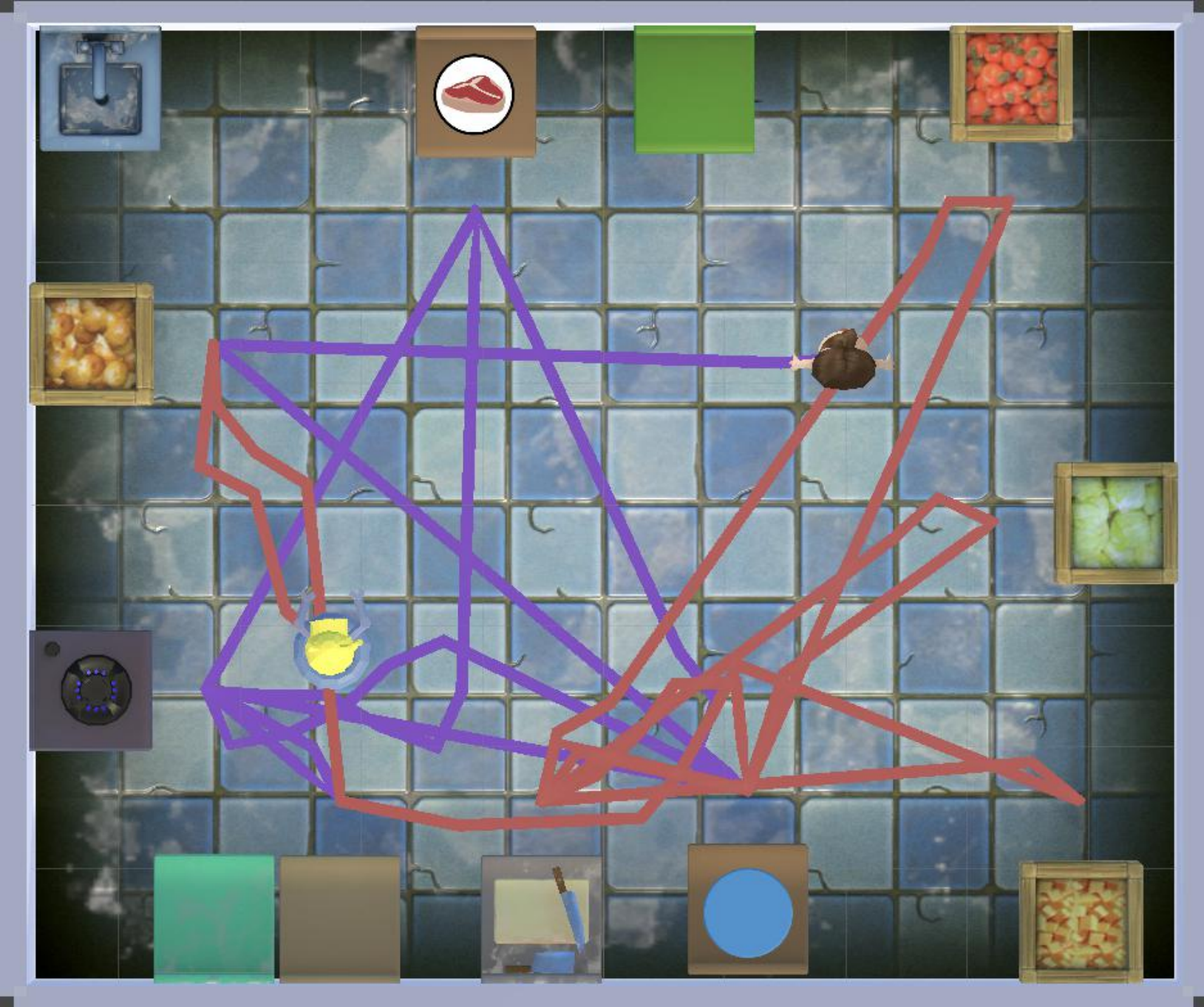}}
{\includegraphics[width = 0.24\textwidth]{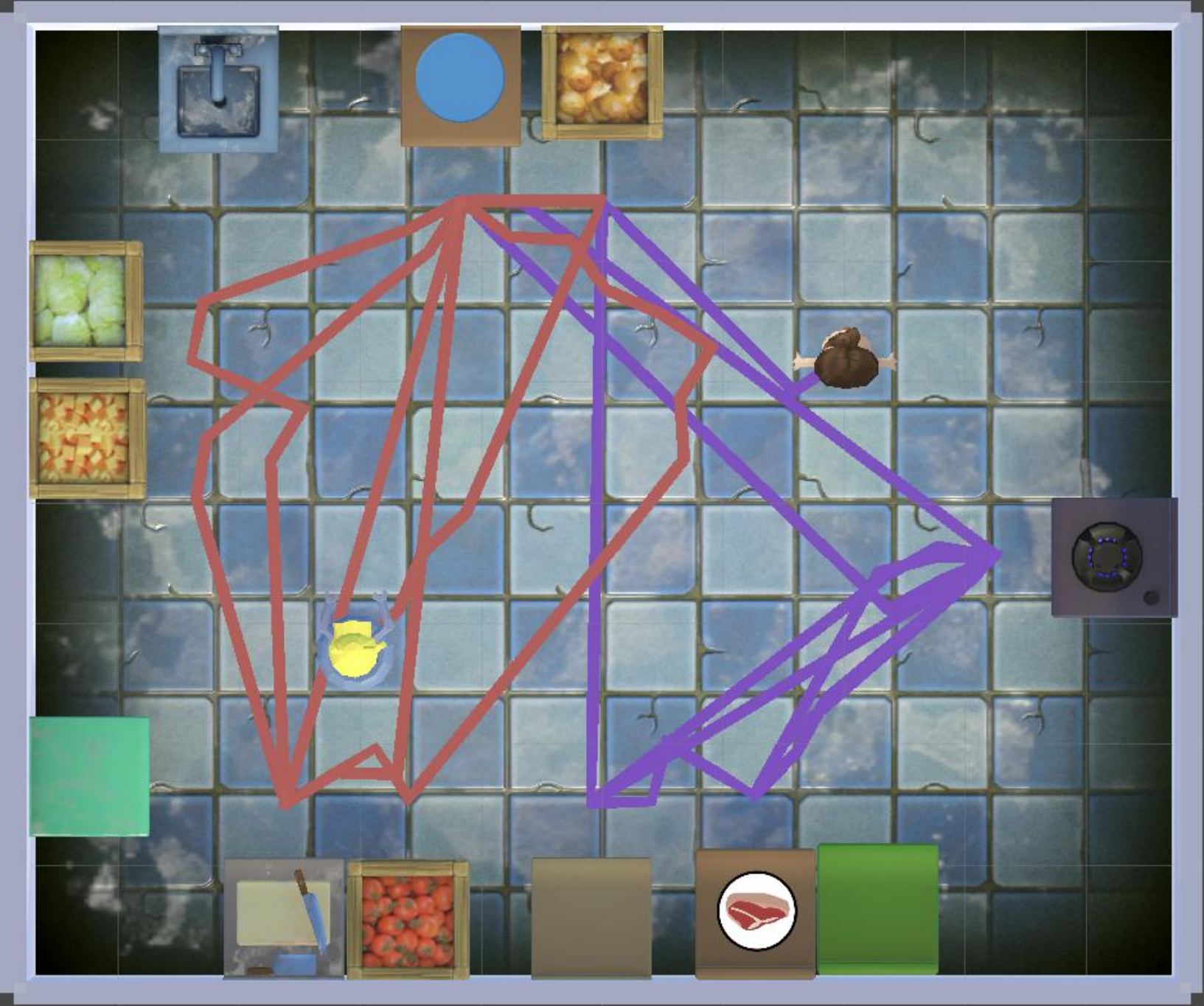}}
\caption{Two kitchen space designs with a smaller area}
\label{fig:small-kitchen}
\end{figure}

Here we use the proposed method to design a smaller kitchen for the same task. The kitchen has a boundary length reduced by 20\%, then we have a 36\% smaller kitchen than the regular room. Fig~\ref{fig:small-kitchen} shows two layouts of the smaller kitchen designed for making two burgers. The designs closely resemble those of the larger kitchen. 

\begin{figure}[]
\centering
{\includegraphics[width = 0.48\textwidth]{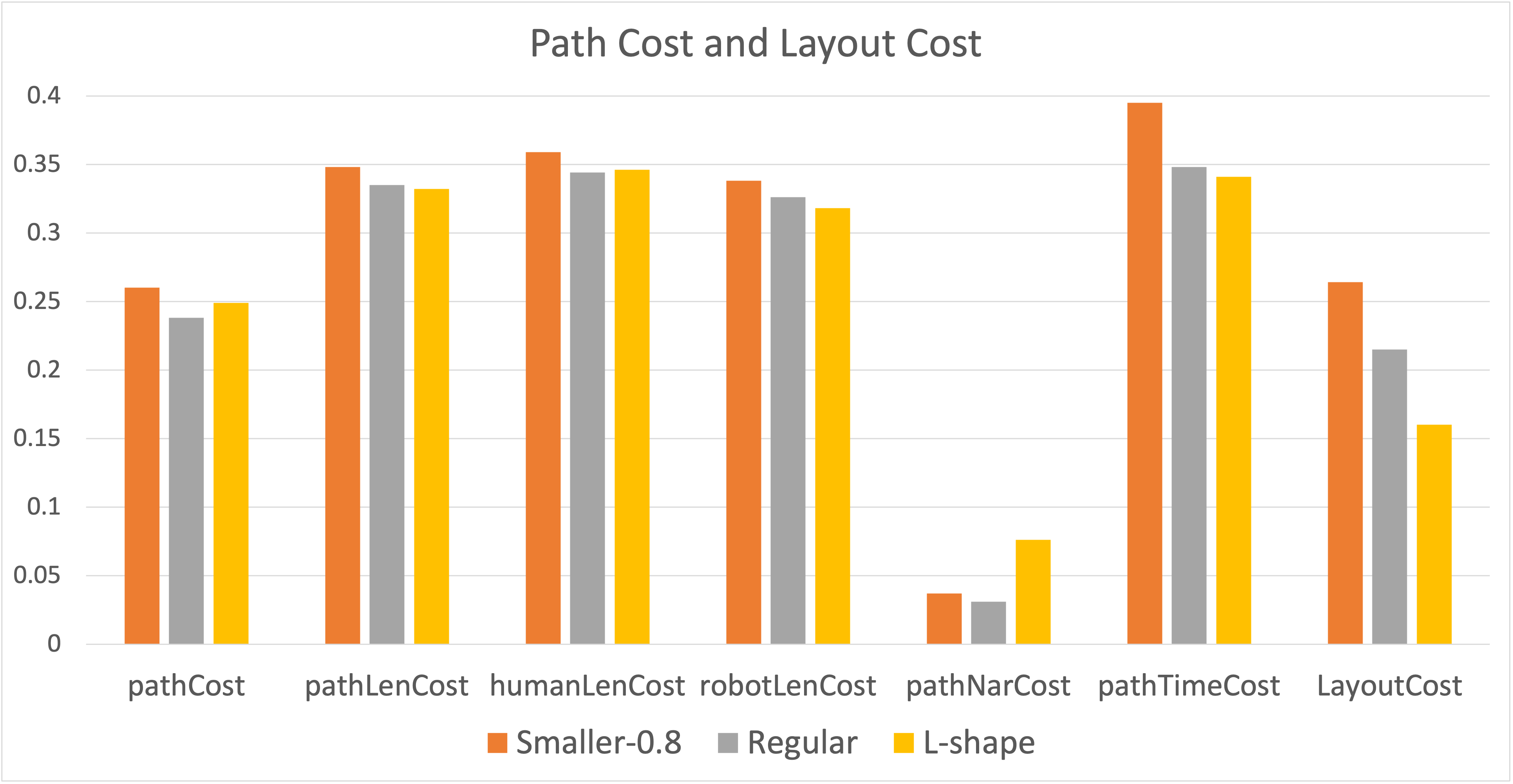}}
\caption{Path costs and layout costs for all various kitchens designed for the same task.  }
\label{fig:spc}
\end{figure}

Fig~\ref{fig:spc} shows the different types of path costs and layout costs in various kitchens with the same task. In every case, the costs of the smaller room are higher than the costs of the regular room, note that we normalize the path cost by room boundary. Finding a high-quality path is harder in a smaller room. 
%However, the path length of a smaller kitchen is lower than the regular room, see Fig~\ref{fig:spl}.

%Fig.~\ref{fig:small} shows two designs of the smaller bedroom. The layouts are similar to those of the larger room. 
%However,
%since the bedroom is more \fa{crowded}\fd{crowed}, the motion planner takes much more time \fa{and is more difficult to converge.} 
%the average path finding time of the proposed method in the small size room is 30\% higher than that of the original bedroom. 

\begin{figure}[]

{\includegraphics[width = 0.24\textwidth]{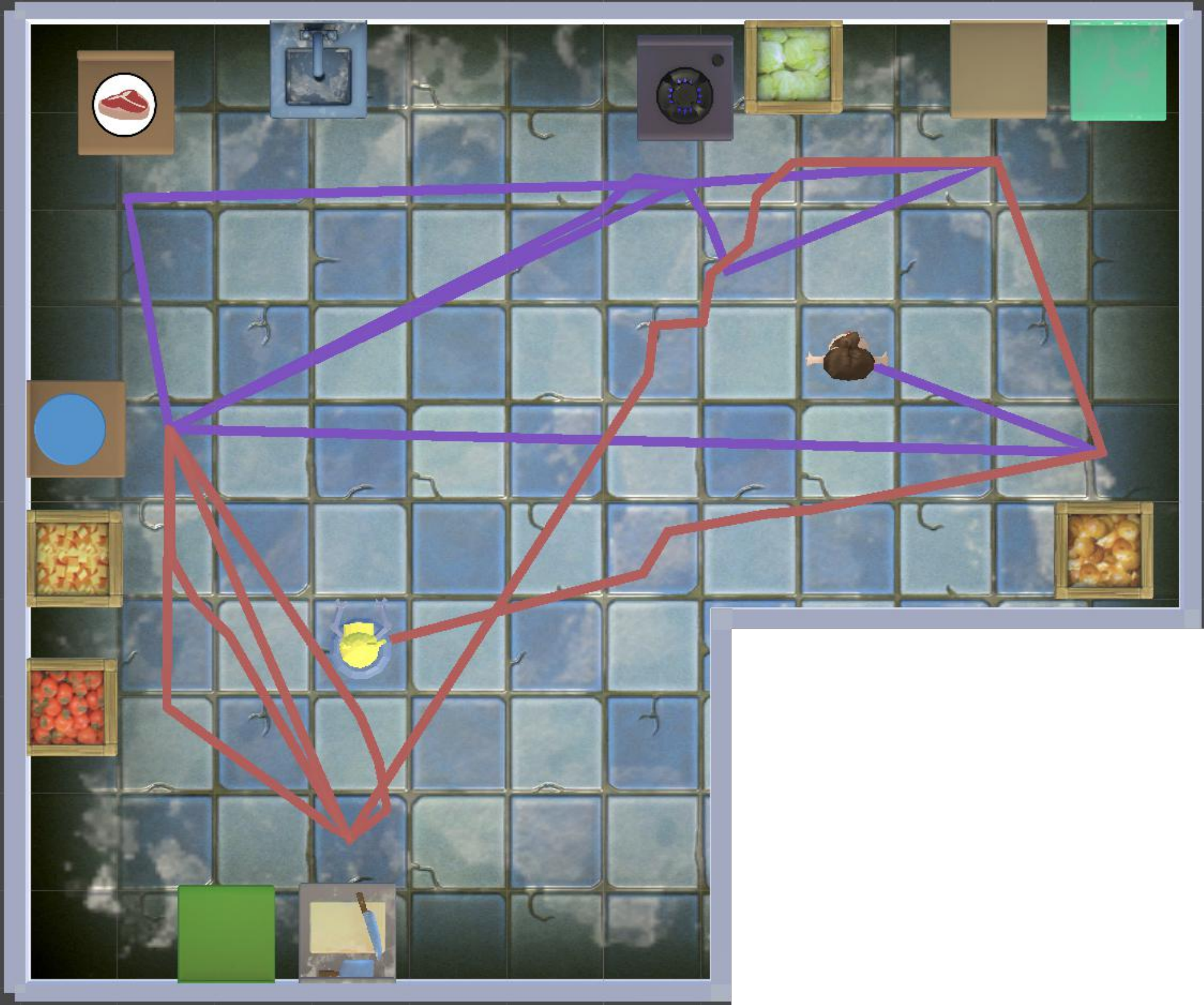}}
{\includegraphics[width = 0.24\textwidth]{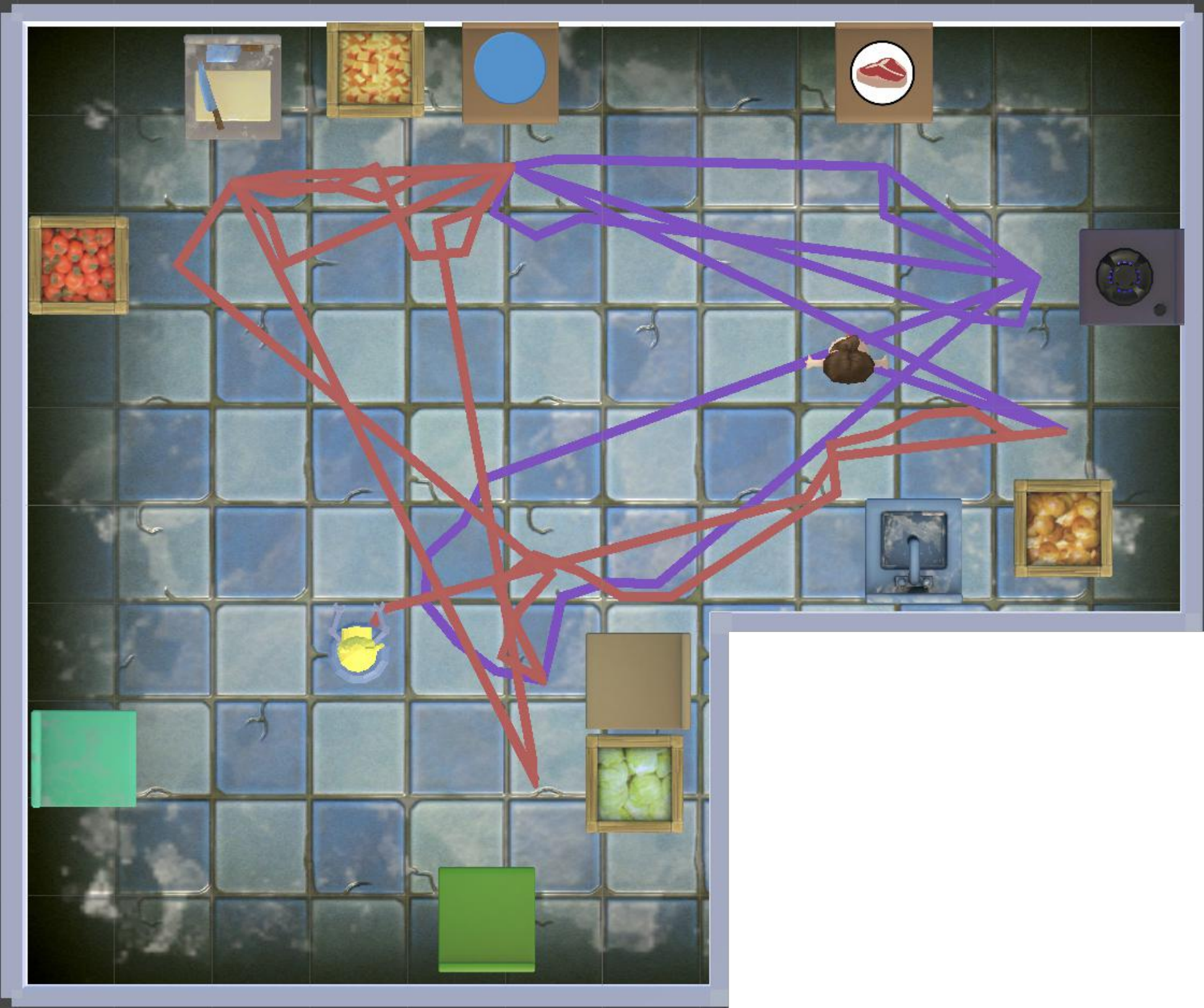}}
\caption{Two designs of L-shape of kitchen}
\label{fig:l-kitchen}
\end{figure}

%\begin{figure}[]
%\centering
%{\includegraphics[width = 0.48\textwidth]{Kfigs/shape_pathlength.png}}

%\caption{Different types of path lengths in various kitchens with the same %task.  }
%\label{fig:spl}
%\end{figure}

%\begin{figure}[t]\centering
%{\includegraphics[width = 0.48\textwidth]{Kfigs/shape_finishtime.png}}
%\caption{Task completion time in various kitchens with the same task. }
%\label{fig:sft}
%\end{figure}

\subsection{Design an L-shaped Kitchen}

We further apply the proposed method to design an L-shaped kitchen using the same task. Fig~\ref{fig:l-kitchen} shows two different layouts of an L-shape of the kitchen when designed for two burgers. 

Fig~\ref{fig:spc} shows that the total path cost of an L-shaped kitchen is higher than the costs of a regular kitchen with the same task, especially for path narrow cost, which is about three times the case of a regular kitchen. This is because the path is more constrained in a non-convex room. %However, the path length of an L-shaped kitchen is smaller than the regular room, see Fig~\ref{fig:spl}.

%We further evaluate the proposed method in a room with more complex boundary using an L-shape living room. Fig.~\ref{fig:Lshape} shows two designs generated for the L-shape living room. Both layouts satisfy the layout and path requirements.

\section{Conclusion}
We developed a computational design method for the kitchen that enhances Human-Robot collaboration. The challenge is the optimization calls the motion planner multiple times during the process and the computational cost of the motion planner is high, so we investigate a decentralized multi-agent motion planner that can coordinate human and robot motion efficiently. 
%and investigate a multi-agent motion planner that can coordinate human and robot  motion by probability inference and reactive planning. 
The results show that the proposed framework can produce practical designs of kitchens that allow humans and robots to be more productive. In addition, The experiments demonstrate that the approach is reliable in generating layouts for more crowded spaces or complex spaces. A major drawback of this work is that the number and the type of recipes are small. When the number of recipes increases, the number of sub-task, and the number of path segments increases, finding a valid multi-agent collision-free path efficiently is quite difficult.

%this is not practical in the real world. In the future, we propose to include multiple recipes in a kitchen space design, such as making a burger, salad, soup, and so on. 

%One major limitation of our work is that the number and type of objects in the room are small. As the number of objects in the room increases, the number of possible sequences and tours increases exponentially. 

\bibliographystyle{IEEEtran}
\bibliography{main}

% Generated by IEEEtran.bst, version: 1.14 (2015/08/26)
\begin{thebibliography}{10}
\providecommand{\url}[1]{#1}
\csname url@samestyle\endcsname
\providecommand{\newblock}{\relax}
\providecommand{\bibinfo}[2]{#2}
\providecommand{\BIBentrySTDinterwordspacing}{\spaceskip=0pt\relax}
\providecommand{\BIBentryALTinterwordstretchfactor}{4}
\providecommand{\BIBentryALTinterwordspacing}{\spaceskip=\fontdimen2\font plus
\BIBentryALTinterwordstretchfactor\fontdimen3\font minus
  \fontdimen4\font\relax}
\providecommand{\BIBforeignlanguage}[2]{{%
\expandafter\ifx\csname l@#1\endcsname\relax
\typeout{** WARNING: IEEEtran.bst: No hyphenation pattern has been}%
\typeout{** loaded for the language `#1'. Using the pattern for}%
\typeout{** the default language instead.}%
\else
\language=\csname l@#1\endcsname
\fi
#2}}
\providecommand{\BIBdecl}{\relax}
\BIBdecl

\bibitem{zhi2021designing}
J.~Zhi, L.-F. Yu, and J.-M. Lien, ``Designing human-robot coexistence space,''
  \emph{IEEE Robotics and Automation Letters}, vol.~6, no.~4, pp. 7161--7168,
  2021.

\bibitem{ghost2016}
G.~T. Games, ``Overcooked,'' 2016.

\bibitem{yu2011make}
L.~F. Yu, S.~K. Yeung, C.~K. Tang, D.~Terzopoulos, T.~F. Chan, and S.~J. Osher,
  ``Make it home: automatic optimization of furniture arrangement,'' \emph{ACM
  Transactions on Graphics (TOG)-Proceedings of ACM SIGGRAPH 2011, v. 30,(4),
  July 2011, article no. 86}, vol.~30, no.~4, 2011.

\bibitem{zhang2021joint}
Y.~Zhang, H.~Huang, E.~Plaku, and L.-F. Yu, ``Joint computational design of
  workspaces and workplans,'' \emph{ACM Transactions on Graphics (TOG)},
  vol.~40, no.~6, pp. 1--16, 2021.

\bibitem{ma2016action}
R.~Ma, H.~Li, C.~Zou, Z.~Liao, X.~Tong, and H.~Zhang, ``Action-driven 3d indoor
  scene evolution,'' \emph{ACM Transactions on Graphics (TOG)}, vol.~35, no.~6,
  pp. 1--13, 2016.

\bibitem{fu2017adaptive}
Q.~Fu, X.~Chen, X.~Wang, S.~Wen, B.~Zhou, and H.~Fu, ``Adaptive synthesis of
  indoor scenes via activity-associated object relation graphs,'' \emph{ACM
  Transactions on Graphics (TOG)}, vol.~36, no.~6, pp. 1--13, 2017.

\bibitem{kampf2017performance}
A.~K{\"a}mpf-Dern and J.~Konkol, ``Performance-oriented office
  environments--framework for effective workspace design and the accompanying
  change processes,'' \emph{Journal of Corporate Real Estate}, 2017.

\bibitem{ribino2018agent}
P.~Ribino, M.~Cossentino, C.~Lodato, and S.~Lopes, ``Agent-based simulation
  study for improving logistic warehouse performance,'' \emph{Journal of
  Simulation}, vol.~12, no.~1, pp. 23--41, 2018.

\bibitem{bauer2008human}
A.~Bauer, D.~Wollherr, and M.~Buss, ``Human--robot collaboration: a survey,''
  \emph{International Journal of Humanoid Robotics}, vol.~5, no.~01, pp.
  47--66, 2008.

\bibitem{alami2005task}
R.~Alami, A.~Clodic, V.~Montreuil, E.~A. Sisbot, and R.~Chatila, ``Task
  planning for human-robot interaction,'' in \emph{Proceedings of the 2005
  joint conference on Smart objects and ambient intelligence: innovative
  context-aware services: usages and technologies}, 2005, pp. 81--85.

\bibitem{carroll2019utility}
M.~Carroll, R.~Shah, M.~K. Ho, T.~Griffiths, S.~Seshia, P.~Abbeel, and
  A.~Dragan, ``On the utility of learning about humans for human-ai
  coordination,'' \emph{Advances in Neural Information Processing Systems},
  vol.~32, pp. 5174--5185, 2019.

\bibitem{balch1994communication}
T.~Balch and R.~C. Arkin, ``Communication in reactive multiagent robotic
  systems,'' \emph{Autonomous robots}, vol.~1, pp. 27--52, 1994.

\bibitem{wu2021too}
S.~A. Wu, R.~E. Wang, J.~A. Evans, J.~B. Tenenbaum, D.~C. Parkes, and
  M.~Kleiman-Weiner, ``Too many cooks: Bayesian inference for coordinating
  multi-agent collaboration,'' \emph{Topics in Cognitive Science}, vol.~13,
  no.~2, pp. 414--432, 2021.

\bibitem{shum2019theory}
M.~Shum, M.~Kleiman-Weiner, M.~L. Littman, and J.~B. Tenenbaum, ``Theory of
  minds: Understanding behavior in groups through inverse planning,'' in
  \emph{Proceedings of the AAAI conference on artificial intelligence},
  vol.~33, no.~01, 2019, pp. 6163--6170.

\bibitem{van2005prioritized}
J.~P. Van Den~Berg and M.~H. Overmars, ``Prioritized motion planning for
  multiple robots,'' in \emph{2005 IEEE/RSJ International Conference on
  Intelligent Robots and Systems}.\hskip 1em plus 0.5em minus 0.4em\relax IEEE,
  2005, pp. 430--435.

\bibitem{colledanchise2018behavior}
M.~Colledanchise and P.~{\"O}gren, \emph{Behavior trees in robotics and AI: An
  introduction}.\hskip 1em plus 0.5em minus 0.4em\relax CRC Press, 2018.

\bibitem{lavalle1998rapidly}
S.~M. LaValle, ``Rapidly-exploring random trees: A new tool for path
  planning,'' 1998.

\bibitem{kuffner2000rrt}
J.~J. Kuffner and S.~M. LaValle, ``Rrt-connect: An efficient approach to
  single-query path planning,'' in \emph{Proceedings 2000 ICRA. Millennium
  Conference. IEEE International Conference on Robotics and Automation.
  Symposia Proceedings (Cat. No. 00CH37065)}, vol.~2.\hskip 1em plus 0.5em
  minus 0.4em\relax IEEE, 2000, pp. 995--1001.

\bibitem{kirkpatrick1983optimization}
S.~Kirkpatrick, C.~D. Gelatt, and M.~P. Vecchi, ``Optimization by simulated
  annealing,'' \emph{science}, vol. 220, no. 4598, pp. 671--680, 1983.

\bibitem{metropolis1953equation}
N.~Metropolis, A.~W. Rosenbluth, M.~N. Rosenbluth, A.~H. Teller, and E.~Teller,
  ``Equation of state calculations by fast computing machines,'' \emph{The
  journal of chemical physics}, vol.~21, no.~6, pp. 1087--1092, 1953.

\bibitem{srinivas1994muiltiobjective}
N.~Srinivas and K.~Deb, ``Muiltiobjective optimization using nondominated
  sorting in genetic algorithms,'' \emph{Evolutionary computation}, vol.~2,
  no.~3, pp. 221--248, 1994.

\end{thebibliography}
%\appendix

\end{document}